\documentclass{naturep}
\usepackage[utf8]{inputenc}
\usepackage{lineno}
\usepackage{setspace}
\usepackage{color}
\usepackage{graphicx}
\usepackage{booktabs}
\usepackage{multirow}
\usepackage{amssymb}
\usepackage{amsmath}
\usepackage{hyperref}
\usepackage{adjustbox}
\usepackage{multicol}
\usepackage{subcaption}
\usepackage{tablefootnote}
\usepackage[margin=0.6in]{geometry}
\usepackage{verbatim}
\usepackage{caption}

\newcommand{\x}{\mathbf{x}}
\newcommand{\z}{\mathbf{z}}
\newcommand{\m}{\mathbf{m}}
\newcommand{\largex}{\mathbf{X}}
\newcommand\Heading[1]{
  \noindent\textbf{\Large{#1}}
}

\newcommand\heading[1]{
  \noindent\textbf{\large{#1}}
}

\newcommand\hheading[1]{
  \noindent\textbf{#1}
}

\title{\raggedright{\textbf{A Foundation Model for Spatial Proteomics}}}
\raggedright{
\author{Muhammad Shaban$^{1, 2, 3, 4,\dag}$, 
Yuzhou Chang$^{5, 6, \dag}$, 
Huaying Qiu$^{5, \ddag}$, 
Yao Yu Yeo$^{5, \ddag}$, 
Andrew H. Song$^{1, 2, 3, 4, \ddag}$, 
Guillaume Jaume$^{1, 2, 3, 4, \ddag}$, 
Yuchen Wang$^{5}$, 
Luca L. Weishaupt$^{1, 2, 3, 4, 7}$, 
Tong Ding$^{1, 2, 3, 4, 8}$, 
Anurag Vaidya$^{1, 2, 3, 4, 7}$, 
Abdallah Lamane$^{1, 2, 3, 4, 7}$, 
Daniel Shao$^{1, 2, 3, 4, 7}$, 
Mohammed Zidane$^{9}$, 
Yunhao Bai$^{4}$, 
Paige McCallum$^{5, 10}$, 
Shuli Luo$^{5}$, 
Wenrui Wu$^{5}$, 
Yang Wang$^{5}$, 
Precious Cramer$^{5}$, 
Chi Ngai Chan$^{5}$, 
Pierre Stephan$^{1, 5}$, 
Johanna Schaffenrath$^{5}$, 
Jia Le Lee$^{5}$, 
Hendrik A. Michel$^{5}$, 
Caiwei Tian$^{1, 2, 3, 4, 11}$, 
Cristina Almagro-Perez$^{1, 2, 3, 4, 7}$, 
Sophia J. Wagner$^{1, 2, 3, 4}$, 
Sharifa Sahai$^{1, 2, 3, 4, 12}$, 
Ming Y. Lu$^{1, 2, 3, 4}$, 
Richard J. Chen$^{1, 2, 3, 4}$, 
Andrew Zhang$^{1, 2, 3, 4, 7}$, 
Mark Edward M. Gonzales$^{13}$, 
Ahmad Makky$^{9}$, 
Jia-Ying Joey Lee$^{13}$, 
Hao Cheng$^{6}$, 
Nourhan El Ahmar$^{1}$, 
Sayed Matar$^{1}$, 
Maximilian Haist$^{14, 15}$, 
Darci Phillips$^{16}$, 
Yuqi Tan$^{14, 15}$, 
Garry P. Nolan$^{14, 15}$, 
W. Richard Burack$^{17}$, 
Jacob D. Estes$^{18}$, 
Jonathan T.C. Liu$^{19}$, 
Toni K. Choueiri$^{20}$, 
Neeraj Agarwal$^{21}$, 
Marc Barry$^{21}$, 
Scott J. Rodig$^{1}$, 
Long Phi Le$^{2}$, 
Georg Gerber$^{1}$, 
Christian M. Schürch$^{22}$, 
Fabian J. Theis$^{23}$, 
Youn H Kim$^{16}$, 
Joe Yeong$^{24}$, 
Sabina Signoretti$^{1}$, 
Brooke E. Howitt$^{15}$, 
Lit-Hsin Loo$^{13}$, 
Qin Ma$^{6, 25}$, 
Sizun Jiang$^{1,4,5,26, \ast}$, 
Faisal Mahmood$^{1, 2, 3, 4, 27, \ast}$
}
\date{}
    
\makeatletter
\let\saved@includegraphics\includegraphics
\AtBeginDocument{\let\includegraphics\saved@includegraphics}
\renewenvironment*{figure}{\@float{figure}}{\end@float}
\makeatother

\begin{document}
\maketitle
\begin{affiliations}
\item Department of Pathology, Brigham and Women's Hospital, Harvard Medical School, Boston, MA, USA
\item Department of Pathology, Massachusetts General Hospital, Harvard Medical School, Boston, MA, USA
\item Data Science Program, Dana-Farber Cancer Institute, Boston, MA, USA
\item Broad Institute of Harvard and MIT, Cambridge, MA, USA
\item Center for Virology and Vaccine Research, Beth Israel Deaconess Medical Center, Harvard Medical School, Boston, MA, USA
\item Pelotonia Institute for Immuno-Oncology, The James Comprehensive Cancer Center, The Ohio State University, Columbus, OH, USA
\item Health Sciences and Technology, Harvard-MIT, Cambridge, MA, USA
\item Harvard John A. Paulson School of Engineering And Applied Sciences, Harvard University, Cambridge, MA, USA
\item Department of Pathology and Neuropathology, University Hospital and Comprehensive Cancer Center Tübingen, Tübingen, Germany
\item Division of Experimental Medicine, McGill University; Montreal, Quebec, Canada
\item Department of Biomedical Informatics, Harvard Medical School, Boston, MA, USA
\item Department of Systems Biology, Harvard University, Cambridge, MA, USA
\item Bioinformatics Institute (BII), Agency for Science, Technology and Research (A*STAR), Singapore, Singapore
\item Department of Microbiology and Immunology, Stanford University, Stanford, CA, USA
\item Department of Pathology, Stanford University, Stanford, CA, USA
\item Department of Dermatology, Stanford University, Stanford, CA, USA
\item Department of Pathology and Laboratory Medicine, University of Rochester Medical Center, Rochester, NY, USA
\item Vaccine \& Gene Therapy Institute, Oregon Health \& Science University, Beaverton, OR, USA
\item Department of Mechanical Engineering, Bioengineering, and Laboratory Medicine \& Pathology, University of Washington, Seattle, WA, USA
\item Lank Center for Genitourinary (GU) Oncology, Dana-Farber Cancer Institute, Boston, MA, USA.
\item Department of Pathology, University of Utah, Salt Lake City, UT; ARUP Institute for Clinical and Experimental Pathology, Salt Lake City, UT, USA
\item Cluster of Excellence iFIT (EXC 2180) “Image-Guided and Functionally Instructed Tumor Therapies”, University of Tübingen, Germany
\item Institute of Computational Biology, Helmholtz Center, Munich, Germany
\item Institute of Molecular and Cell Biology (IMCB), Agency for Science Technology and Research (A*STAR), Singapore, Singapore
\item Department of Biomedical Informatics, College of Medicine, Ohio State University, Columbus, OH, USA
\item Department of Pathology, Dana-Farber Cancer Institute, Boston, MA, USA
\item Harvard Data Science Initiative, Harvard University, Cambridge, MA, USA
\item[$^{\dag}$] Contributed equally (Co-first authorship)
\item[$^{\ddag}$] Contributed equally (Co-second authorship)
\item[$^{\ast}$] \textbf{Corresponding authors}: Sizun Jiang (sjiang3@bidmc.harvard.edu) and Faisal Mahmood (faisalmahmood@bwh.harvard.edu)
 \end{affiliations}

\newpage
\begin{abstract}
\begin{spacing}{1.2}
\Heading{Abstract}

Foundation models have begun to transform image analysis by acting as pretrained generalist backbones that can be adapted to many tasks even when post-training data are limited, yet their impact on spatial proteomics, imaging that maps proteins at single-cell resolution, remains limited. Here, we introduce KRONOS, a foundation model built for spatial proteomics. KRONOS was trained in a self-supervised manner on over 47 million image patches covering 175 protein markers, 16 tissue types, and 8 fluorescence-based imaging platforms. We introduce key architectural adaptations to address the high-dimensional, multi-channel, and heterogeneous nature of multiplex imaging. We demonstrate that KRONOS learns biologically meaningful representations across multiple scales, ranging from cellular and microenvironment to tissue levels, enabling it to address diverse downstream tasks, including cell phenotyping, region classification, and patient stratification. Evaluated across 11 independent cohorts, KRONOS achieves state-of-the-art performance across cell phenotyping, treatment response prediction, and retrieval tasks, and is highly data-efficient. KRONOS also introduces the paradigm of segmentation-free patch-level processing for efficient and scalable spatial proteomics analysis, allowing cross-institutional comparisons, and as an image reverse search engine for spatial patterns.
Together, these results position KRONOS as a flexible and scalable tool for spatial proteomics. The model is publicly accessible at \url{https://github.com/mahmoodlab/KRONOS}.
\end{spacing}

\end{abstract}

\newpage
\begin{spacing}{1.5}
\Heading{Introduction}

Spatial proteomics is a rapidly advancing class of imaging technologies that measures the spatial distribution of protein expression at single-cell resolution within intact tissue sections\cite{karimi2024method}. Unlike traditional immunohistochemistry, spatial proteomics can capture tens to over a hundred protein markers simultaneously in the same tissue sample\cite{desouza2024multiplex}. This multiplexing capacity enables profiling of cell types, functional states, and signaling pathways within their native spatial context. As a result, spatial proteomics has become essential for quantifying immune infiltration, identifying rare cell populations, resolving tumor–stroma boundaries, and mapping spatial patterns of therapeutic targets, all key determinants of disease progression and treatment response\cite{lundberg2019spatial, lewis2021spatial, wu2022graph, de2024multiplex, yeo2024hitchhiker}.

\noindent Standard spatial proteomics pipelines typically rely on cell segmentation followed by single-marker thresholding or rule-based mechanisms to assign cell phenotypes\cite{jackson2020single, schurch2020coordinated, yeo2024same, zhu2025multi, yeo2024epstein, lin2023multiplexed, keren2018structured, jiang2022combined}. While effective for identifying canonical cell types, these approaches treat each marker independently and reduce multiplexed profiles to coarse categories. Moreover, this overlooks cell states characterized by the coordinated expression of multiple markers, which can be continuous and influenced by spatial factors. Moreover, segmentation-based methods struggle in crowded, artifact-prone, or morphologically complex regions. 

\begin{figure}[!t]
\centering
\includegraphics[width=1.00\textwidth]{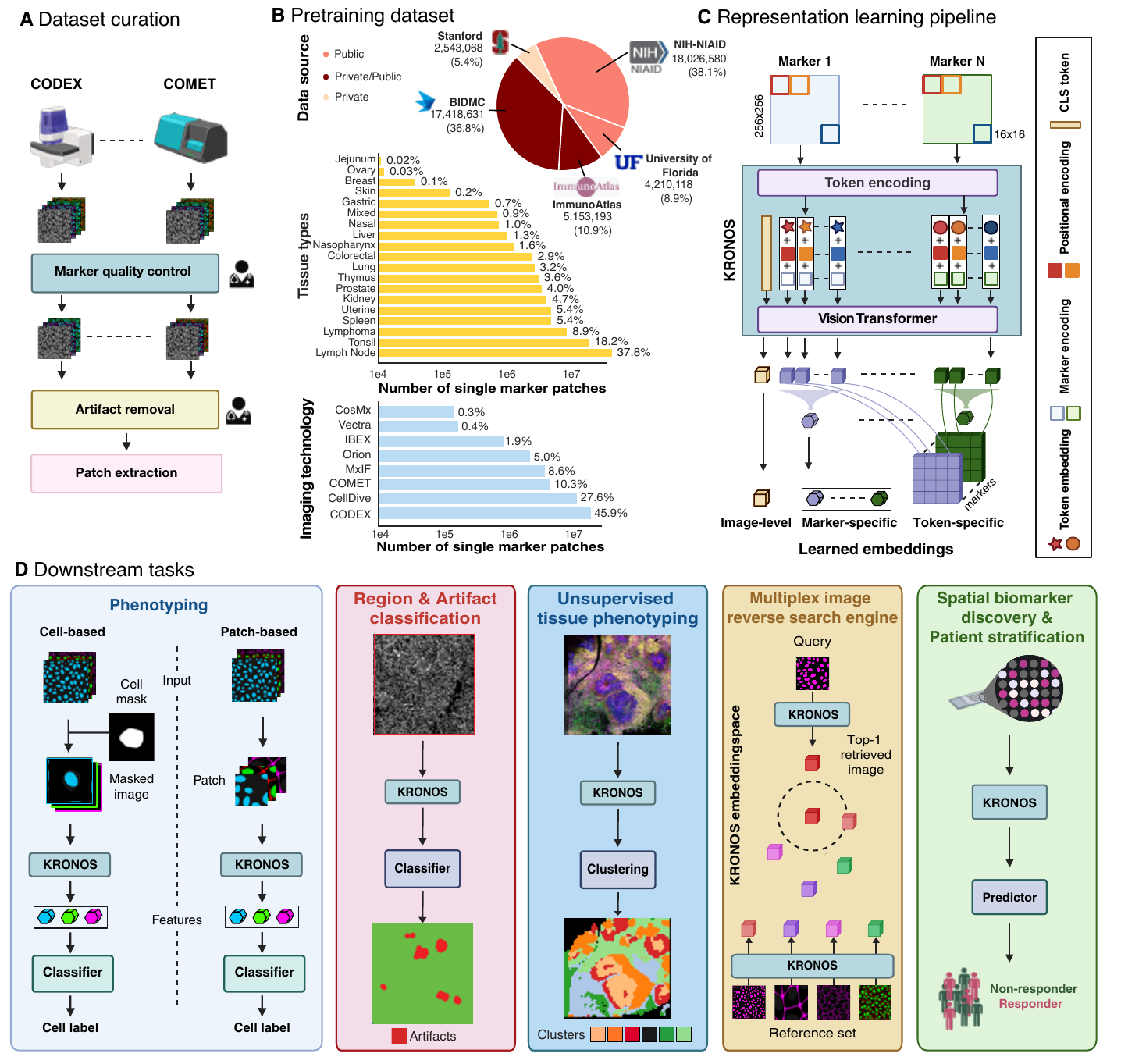}
\caption{\textbf{Overview of KRONOS: A foundation model for multiplex spatial proteomics.} 
\textbf{A}, KRONOS was pretrained on SPM-47M, a collection of multiplexed image sources from 30 cohorts and eight fluorescence-based imaging platforms.
\textbf{B}, SPM-47M comprises over 47 million single-marker image patches derived from 175 unique protein markers and 16 distinct tissue types. 
\textbf{C}, KRONOS is built on a Vision Transformer, DINO-v2, adapted to multiplexed spatial proteomics. 
\textbf{D}, KRONOS was evaluated across a range of downstream tasks, including cell phenotyping, region classification, artifact detection, unsupervised tissue phenotyping, cross-scale image retrieval akin to an image search engine, and patient stratification.
Detailed illustrations on KRONOS pretraining and workflows can be found in \textbf{Supplementary Figure~\ref{fig:kronos}} and \textbf{~\ref{fig:kronos_workflow}}, respectively. 
}
\label{fig:overview}
\end{figure}

We argue that a foundation model\cite{song2023artificial, uni, giga-path, virchow, prism, phikon, phikon-2, chief, titan, biomedparse, microsnoop} trained directly on diverse and extensive amounts of unlabeled multiplexed images can overcome these limitations. Here, we introduce KRONOS, a general-purpose foundation model for spatial proteomics. Trained on a curated corpus of over 47 million image patches encompassing 16 tissue types, eight fluorescence-based imaging technologies, and 175 protein markers (\textbf{Figure \ref{fig:overview}A,B}), KRONOS employs a Vision Transformer (ViT) architecture optimized for multiplexed imaging (\textbf{Figure \ref{fig:overview}C} and \textbf{Supplementary Figure \ref{fig:kronos}}). By encoding all multiplex markers jointly and learning from dense multi-channel image patches in a self-supervised fashion\cite{dinov2}, KRONOS can generate general-purpose representations that capture both fine-grained phenotypes and larger tissue context.

We demonstrate that KRONOS delivers robust, generalizable performance across several downstream tasks, including cell phenotyping, region classification, artifact detection, and downstream analyses, such as the ability to link spatial features with clinical readouts and ``reverse search'' of tissue microenvironments across studies, all while maintaining high interpretability and data efficiency (\textbf{Figure \ref{fig:overview}D}). Our results establish KRONOS as a foundation model for spatial proteomics, capable of exploring the latent structure within multiplexed tissue images and enabling scalable, annotation-efficient biological discovery.

\clearpage
\Heading{Results}

\heading{Foundation model pretraining on multiplexed spatial proteomics}

The ability of a foundation model to learn powerful and robust representations depends critically on the diversity, scale, biological fidelity, and quality of its pretraining dataset\cite{dinov2, uni}. In contrast to common clinically deployed imaging modalities such as hematoxylin and eosin or single-plex immunohistochemistry, multiplexed spatial proteomics data remain scarce and tailored to specific problem statements, largely due to their costs and technical limitations in implementation at scale\cite{ibex, lymphoma, hubmap, lee2021627, yeotumor}. To enable large-scale pretraining of KRONOS, we curated SPM-47M, a collection of 47 million single-marker image patches sourced from 30 distinct data sources. SPM-47M integrates multiplexed images from both public (56\%, e.g., ImmunoAtlas\cite{lee2021627} among others\cite{ibex, lymphoma, hubmap}) and in-house (44\%) cohorts, spanning eight distinct fluorescence-based spatial proteomics platforms, including CODEX\cite{goltsev2018codex}, MxIF\cite{parra2017mxif}, COMET\cite{rivest2023comet}, IBEX\cite{ibex} (\textbf{Figure \ref{fig:overview}A}), and covering 16 tissue types ranging from normal to malignant conditions (\textbf{Figure \ref{fig:overview}B, Supplementary Table \ref{tab:pretrain_dataset}}). 

To ensure data quality, we implemented a standardized preprocessing pipeline, where each protein marker underwent manual quality assessment to evaluate staining fidelity and remove channels with non-specific or poor signal. Tissue masking was applied to isolate valid tissue regions while excluding artifacts from imaging or slide preparation (e.g., saturation of signal, blurring, tissue folding). Processed tissue images were then divided into image patches of size $256\times256$, resulting in 3.67 million multiplex image patches and over 47 million single-marker image patches. Overall, SPM-47M spans 175 distinct markers, with each multiplexed image containing between 3 and 58 markers (\textbf{Supplementary Table \ref{tab:marker_list}}). 

Due to the differences in study designs, imaging platforms, and panel compositions, the number and identity of markers vary widely across studies. This renders current foundation models for vision, optimized for three-channel RGB inputs, not directly applicable to multi-channel images\cite{vit, mae, dino, ibot, dinov2, i-jepa}. To deal with the variable number of markers, previous works have extended three-channel ViTs to $N$-channel models\cite{channel-vit, chada-vit, scdino, DiChaViT, doron2023unbiased, lu2019learning} on multiplex cell profiling datasets\cite{sc-hpa, wtc-11, cellpainting, channel-vit, cytoself, chammi}. However, existing models remain limited to a fixed number of channels and do not scale beyond $N$-channels per image. Channel-agnostic models have also been evaluated, where either no channel-specific embeddings were used \cite{ca-mae, kenyon2024vitally} or marker embeddings were generated from a pretrained protein language model\cite{wenckstern2025ai}. These models use masked autoencoding\cite{he2022masked}, pretraining on fluorescent cell profiling datasets and image mass cytometry\cite{giesen2014imc} datasets, respectively. 

KRONOS utilizes a combination of small image tokens, spatial position encoding, and marker identity encoding to learn marker-specific and spatially contextualized features from the input marker panel (\textbf{Figure \ref{fig:overview}C}, \textbf{Supplementary Figure \ref{fig:kronos}}). In contrast to other in-domain foundation models\cite{ca-mae, kenyon2024vitally, wenckstern2025ai}, KRONOS is based on the ViT small architecture pretrained on SPM-47M, with DINO-v2\cite{dinov2}, a self-supervised learning pretraining strategy that has been highly successful when applied to histology imaging to date\cite{uni, giga-path, virchow, phikon}.

\begin{figure*}
\centering
\includegraphics[width=0.90\textwidth]{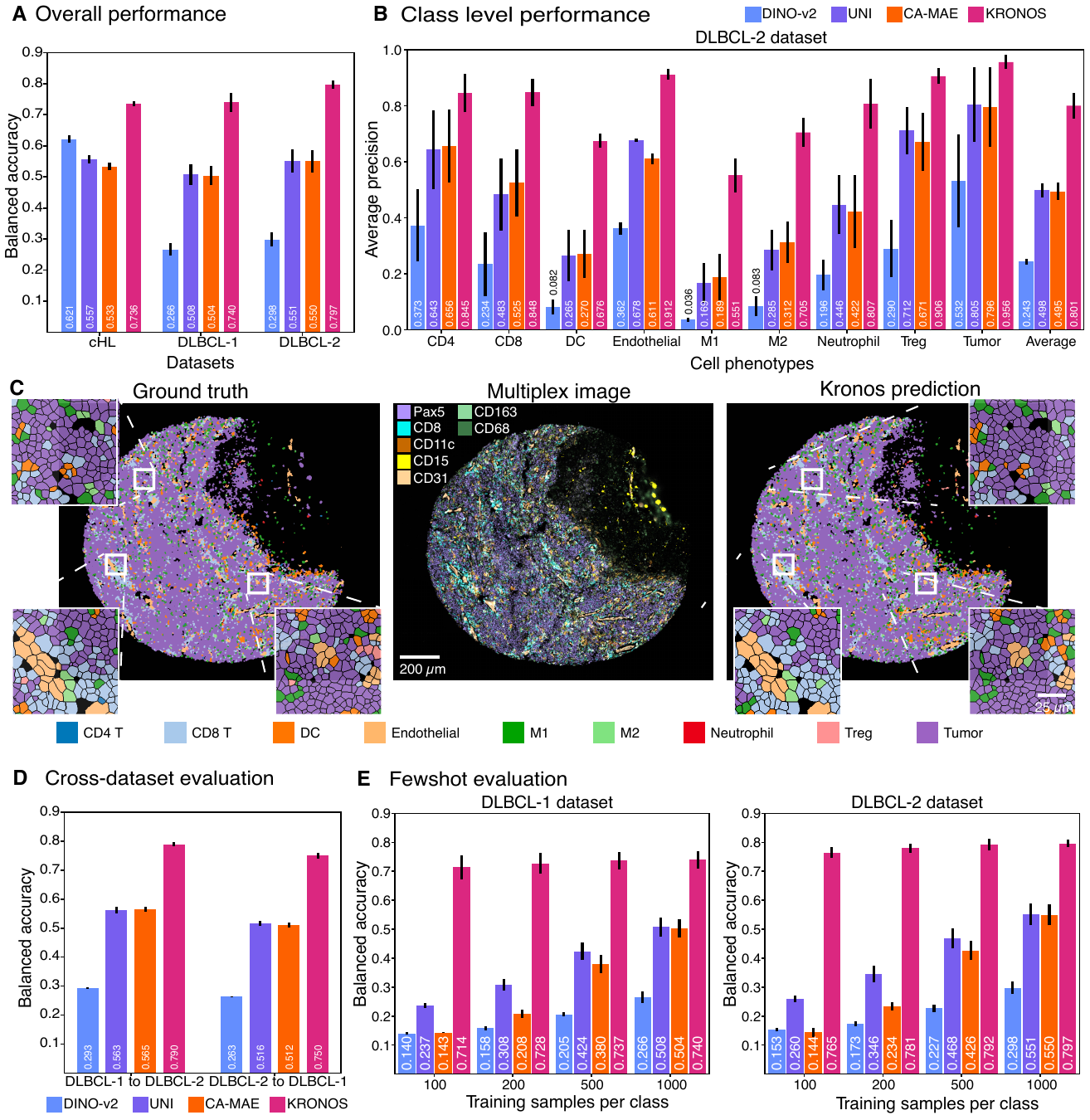}
\caption{\textbf{Benchmarking KRONOS for cell phenotyping across three spatial proteomics datasets using linear probing evaluation.} 
\textbf{A}, Overall performance comparison across three spatial proteomics datasets: 16-class classical Hodgkin lymphoma (cHL, n=134,552 cells) and two 9-class diffuse large B cell lymphoma datasets (DLBCL-1, n=361,341 cells and DLBCL-2, n=1,098,443 cells). 
\textbf{B}, Class-wise average precision comparison of KRONOS and baselines on the DLBCL-2 cohort. 
\textbf{C}, Visual comparison of ground truth (left) and KRONOS-predicted (right) cell phenotype maps, alongside the corresponding multiplexed image (center) annotated with relevant markers. 
\textbf{D}, Cross-dataset generalization performance of cell phenotyping trained on one DLBCL dataset and evaluated on the other. 
\textbf{E}, Few-shot evaluation of KRONOS and baselines trained using 100 to 1,000 samples per class on DLBCL-1 and DLBCL-2. In \textbf{A,B,D,E}, each bar represents the mean performance, where a higher score indicates better performance, with error bars indicating the standard deviation across four folds. M1 and M2 denote two macrophage phenotypes. More results can be found in \textbf{Supplementary Figure~\ref{fig:supp_image_classification}}.
}
\label{fig:cell_phenotyping}    
\end{figure*}

\heading{Cell phenotyping with KRONOS}

To evaluate the quality and biological relevance of representations learned by KRONOS, we assessed its performance on cell phenotyping, a fundamental task in spatial proteomics with downstream implications for functional analysis, biomarker discovery, patient risk stratification, and therapy response prediction\cite{jackson2020single, magness2024deep, shaban2024maps} (\textbf{Supplementary Figure \ref{fig:kronos_workflow}}). 

\noindent\textbf{Setup and baselines.} We benchmarked KRONOS on a cohort of patients diagnosed with classical Hodgkin's Lymphoma\cite{shaban2024maps} (cHL, 16-class classification task and over 134,000 annotated cells), and two cohorts of patients with Diffuse large B-cell lymphoma\cite{yeo2024same}: DLBCL-1 (9-class classification with more than 361,000 cells annotated) and DLBCL-2 (9-class classification with more than 1 million annotated cells), all of which were imaged with CODEX\cite{goltsev2018codex}. Cohorts differ in cell types and marker composition (\textbf{Supplementary Table \ref{tab:cell_dataset}}). All embeddings were extracted from a nuclear stain (DAPI) and phenotypical markers (\textbf{Supplementary Table \ref{tab:phenotyping_markers}}). We first evaluated KRONOS embeddings by performing multi-class classification via four-fold cross-validation. Specifically for each training fold, we randomly selected 1,000 cells per phenotype, segmented each cell with Mesmer\cite{tissuenet}, extracted marker features with KRONOS, and then trained a linear classifier for cell type prediction (\textbf{Supplementary Table \ref{tab:cell_folds}}). We compared KRONOS with two state-of-the-art out-of-domain vision foundation models: DINO-v2\cite{dinov2}, trained on 142M natural images, and UNI\cite{uni}, trained on 100 million H\&E-stained histopathology patches. Furthermore, we compared KRONOS with a domain-specific model (CA-MAE\cite{ca-mae}) trained on multi-channel fluorescent cell profiling images\cite{fay2023rxrx3}. More details can be found in \textbf{Online Methods}, section \textbf{KRONOS evaluation}.

\noindent\textbf{Performance comparison.} Across all three datasets, KRONOS consistently outperformed all baselines, demonstrating superior generalization capacity (\textbf{Figure \ref{fig:cell_phenotyping}A}, \textbf{Supplementary Table  \ref{tab:cell_results}}). In the cHL dataset, KRONOS achieved the highest balanced accuracy of 0.7358 $\pm$ 0.0089, surpassing DINO-v2 (0.6210 $\pm$ 0.0121), UNI (0.5570 $\pm$ 0.0136), and CA-MAE (0.5331 $\pm$ 0.0123). Similar trends were observed in the DLBCL datasets. In DLBCL-1, KRONOS reached a balanced accuracy of 0.7402 $\pm$ 0.0309 (9-class classification), outperforming UNI (0.5077 $\pm$ 0.0333), CA-MAE (0.5041 $\pm$ 0.0314), and DINO-v2 (0.2664 $\pm$ 0.0201). In DLBCL-2, KRONOS achieved a balanced accuracy of 0.7969 $\pm$ 0.0125 (9-class classification), over UNI (0.5511 $\pm$ 0.0377), CA-MAE (0.5503 $\pm$ 0.0368), and DINO-v2 (0.2980 $\pm$ 0.0226). The lower performance of DINO-v2 on the DLBCL datasets relative to cHL may reflect the greater biological and technical heterogeneity present in DLBCL, which comprises tissues from multiple patients, as opposed to the more uniform structure of the single large tissue section in the cHL dataset\cite{shaban2023maps, shaban2024maps}. 

In addition, KRONOS achieved higher precision for all individual cell types across all datasets (\textbf{Figure \ref{fig:cell_phenotyping}B}, \textbf{Supplementary Figure \ref{fig:supp_image_classification}A}, \textbf{Supplementary Table \ref{tab:cell_class_cHL}, \ref{tab:cell_class_dlbcl_1}, and \ref{tab:cell_class_dlbcl_2}}). In the cHL dataset, KRONOS demonstrated high precision in tumor cells (0.9267 $\pm$ 0.0230), CD8+ T cells (0.9089 $\pm$ 0.0141), and lymphatic cells (0.9053 $\pm$ 0.0252), with an average precision of 0.7614 $\pm$ 0.0084, surpassing DINO-v2 (0.6217 $\pm$ 0.0119), UNI (0.5348 $\pm$ 0.0136), and CA-MAE (0.4950 $\pm$ 0.0181). On DLBCL-1, KRONOS exhibits strong performance particularly for endothelial cells (0.9389 $\pm$ 0.0164), tumor cells (0.9700 $\pm$ 0.0094), and CD8+ T cells (0.8721 $\pm$ 0.0256), with an average precision of 0.7567 $\pm$ 0.0392, outperforming UNI (0.4584 $\pm$ 0.0530), CA-MAE (0.4518 $\pm$ 0.0472), and DINO-v2 (0.2227 $\pm$ 0.0229). In DLBCL-2, we observe similar trends as in DLBCL-1, with the overall average precision for KRONOS achieving 0.8007 $\pm$ 0.0462, exceeding UNI (0.4985 $\pm$ 0.0244), CA-MAE (0.4946 $\pm$ 0.0300), and DINO-v2 (0.2432 $\pm$ 0.0103). Visual assessment of the ground truth cell type annotations, along with the multiplex image and KRONOS prediction, confirmed the robustness of the results (\textbf{Figure \ref{fig:cell_phenotyping}C}).

\noindent\textbf{Generalization of KRONOS across cohorts.} Variability in marker composition, tissue type, and disease state makes spatial proteomics datasets inherently heterogeneous. Even within single-disease cohorts, substantial variability can result from batch effects introduced during sample preparation, fixation, staining, and imaging across different platforms or tissue collection and processing sites. To evaluate the generalizability of KRONOS, we conducted cross-dataset transfer experiments using two independent DLBCL datasets that share the same cell type annotations, marker panels, and imaging technology (CODEX), but differ in patient cohorts and collection sites\cite{dlbcl}. Specifically, we trained linear classifiers for cell type classification using 4-fold cross-validation on one dataset (1,000 training cells per class) and evaluated them on the other without fine-tuning, enabling assessment of the transferability of learned representations across protocol differences.

KRONOS consistently outperformed all three baselines (\textbf{Figure \ref{fig:cell_phenotyping}D}, \textbf{Supplementary Table \ref{tab:cross_dataset}}). When trained using DLBCL-1 and tested on DLBCL-2, KRONOS achieves a balanced accuracy of 0.7896 $\pm$ 0.0072, outperforming CA-MAE (0.5651 $\pm$ 0.0074), UNI (0.5629 $\pm$ 0.0108), and DINO-v2 (0.2928 $\pm$ 0.0022). This pattern was maintained in the reverse experiment: training in DLBCL-2 and testing in DLBCL-1, KRONOS again achieved the highest balanced accuracy (0.7505 $\pm$ 0.0100), outperforming CA-MAE (0.5116 $\pm$ 0.0086), UNI (0.5164 $\pm$ 0.0074), and DINO-v2 (0.2632 $\pm$ 0.0021). These results underscore the robustness and transferability of KRONOS representations across independent spatial proteomics datasets with batch variation.

\heading{Label efficiency of KRONOS}

Label efficiency is essential in spatial proteomics, where annotations are expensive and challenging to reuse\cite{jackson2020single, schurch2020coordinated, amitay2023cellsighter}. Differences in tissue types, marker panels, and imaging protocols result in study-specific individualized labels, and expert cell phenotyping is slow and resource-intensive. These constraints limit the scalability of supervised methods. We thus evaluated next the label efficiency of KRONOS in three settings: 1) few-shot phenotyping, 2) human-guided annotation, and 3) zero-shot inference.

\noindent\textbf{Few-shot cell phenotyping.} We assess KRONOS label efficiency by training cell phenotyping classifiers on progressively smaller training sets of 1000, 500, 200, and 100 examples per class. 
Evaluating on cHL, DLBCL-1, and DLBCL-2, we observe that KRONOS consistently outperforms all baselines at every sample size across all datasets 
(\textbf{Figure \ref{fig:cell_phenotyping}E}, \textbf{Supplementary Figure~\ref{fig:supp_image_classification}B}, \textbf{Supplementary Table \ref{tab:limited_dataset_chl}, \ref{tab:limited_dataset_dlbcl1}, \ref{tab:limited_dataset_dlbcl2}}). 
In the cHL dataset, KRONOS achieved a balanced accuracy of 0.6735 $\pm$ 0.0216 with just 100 samples per cell type, increasing to 0.7358 $\pm$ 0.0089 at 1,000 samples. In DLBCL-1, KRONOS maintained high performance with limited supervision, achieving 0.7143 $\pm$ 0.0410 with 100 samples and 0.7402 $\pm$ 0.0309 at 1,000 samples. Similarly, in the DLBCL-2 dataset, KRONOS maintained high performance with limited supervision, achieving 0.7651 $\pm$ 0.0194 with 100 samples and 0.7969 $\pm$ 0.0125 with 1,000 samples.  Notably, KRONOS, even with only 100 samples, significantly outperformed other baselines with ten times more training samples, highlighting the strong data efficiency of KRONOS.

\noindent\textbf{Human-guided cell phenotyping.} Besides clustering-based approaches for computationally stratifying cell phenotypes based on their combined marker expression profiles, another common annotation approach involves graphical user interface-based tools such as QuPath, HALO, and InForm\cite{zhang2025comparison}. Here, an expert labels only a few representative cells for a given image. We tested the ability of KRONOS to use these annotations for training an image-specific classifier to predict cell phenotypes on the rest of the image. In contrast to the previous few-shot setting, all train and test samples are confined to the same image, reducing inter-image variability. We experimented on the DLBCL-1 dataset, where, for each image, few-shot models were trained using randomly sampled 1, 5, 10, and 20 cells per class, and the model performance was evaluated on the remaining cells within the same image. The experiment was repeated 100 times to reduce sampling variability, and the performance was averaged across all images in DLBCL-1. 

KRONOS demonstrated strong performance even under extreme supervision limitations (\textbf{Supplementary Figure~\ref{fig:supp_image_classification}C}, \textbf{Supplementary Table \ref{tab:few_shots}}). With just one labeled cell per class, KRONOS achieved a balanced accuracy of 0.3977 $\pm$ 0.1328, significantly outperforming CA-MAE (0.1363 $\pm$ 0.0564), UNI (0.1365 $\pm$ 0.0630), and DINO-v2 (0.1336 $\pm$ 0.0566). Performance improved markedly with additional samples, reaching 0.6523 $\pm$ 0.0845 with five, 0.7164 $\pm$ 0.0675 with 10, and 0.7534 $\pm$ 0.0581 with 20 labeled samples per phenotype class. In contrast, the improvements with increased samples were minimal for CA-MAE, UNI, and DINO-v2, with performance near random across all numbers of examples. The robust performance of KRONOS in real-world scenarios, where each image contains only a handful of labeled examples, supports the potential of KRONOS for accelerating expert-in-the-loop labeling.

\noindent\textbf{Zero-shot cell phenotyping.} Finally, we aimed to evaluate KRONOS in the extreme setting of zero-shot evaluation, where no supervisory labels are provided. To this end, we perform K-means clustering on the cell embedding space with the number of clusters set to the number of cell types (16 for cHL and 9 for DLBCL-1, DLBCL-2). We then assessed the clustering quality with the adjusted rand index (ARI) metric. Higher ARI indicates that the clusters in the embedding space show clearer separation amongst them and align better with the underlying cell types. We observed that KRONOS reaches the best ARI across all three datasets (\textbf{Supplementary Figure~\ref{fig:supp_image_classification}D}, \textbf{Supplementary Table~\ref{tab:zero_shot}}), achieving 0.3186 $\pm$ 0.0244 (cHL), 0.2589 $\pm$ 0.0344 (DLBCL-1), and 0.2876 $\pm$ 0.0686 (DLBCL-2). The poor performance of the three baseline models, as reflected by their low evaluation metrics, suggests that their embedding spaces failed to capture biologically relevant structure, thereby limiting their ability to phenotype cells.

\begin{figure*}
\centering
\includegraphics[width=0.92\textwidth]{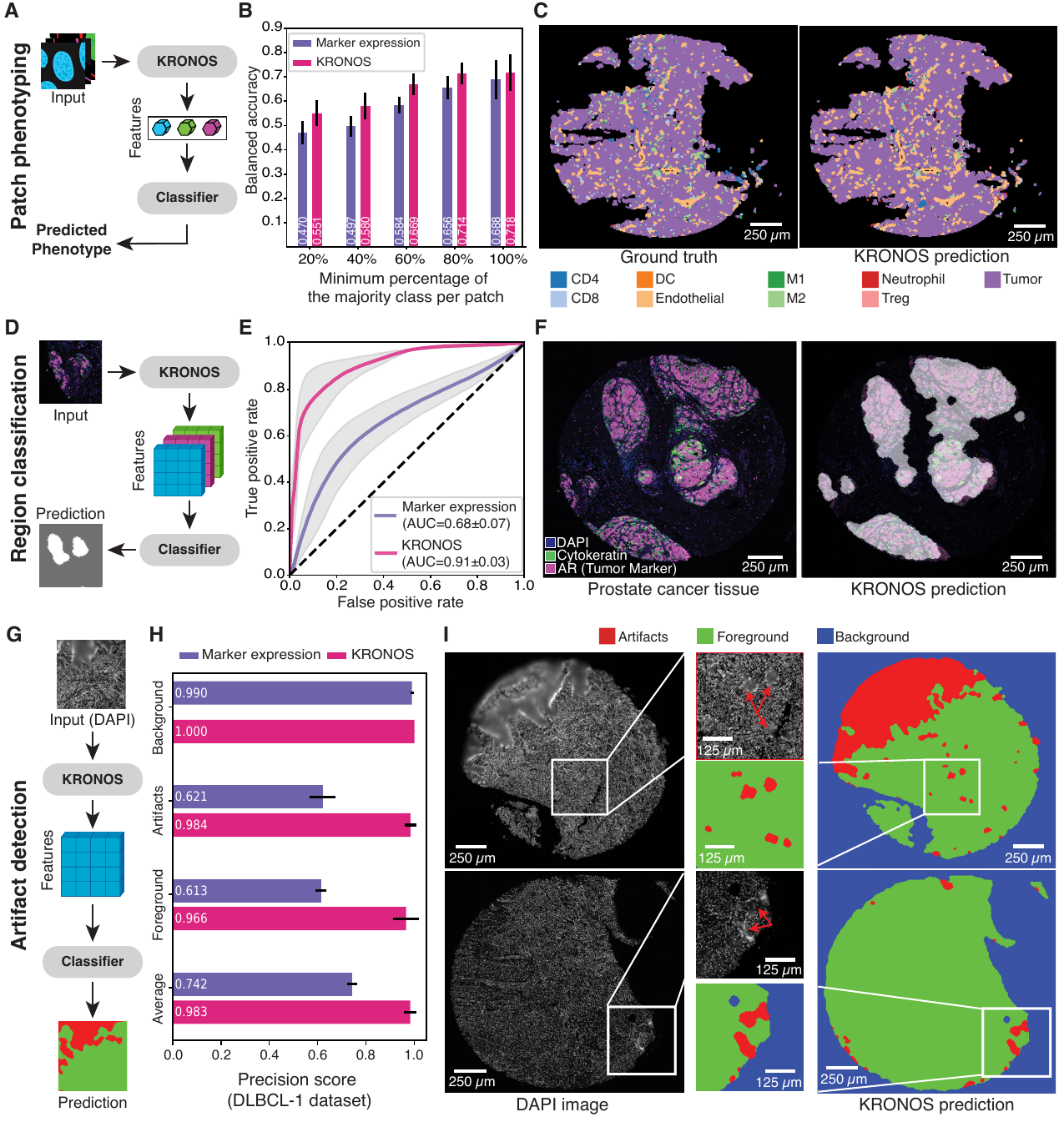}
\caption{\textbf{Comparison of KRONOS against marker expression embeddings for patch phenotyping, region classification, and artifact detection.}  
\textbf{A,B,C} KRONOS for segmentation-free patch phenotyping. 
\textbf{B}: Patch phenotyping performance comparison on the DLBCL-1 dataset. Patch labels were assigned based on the majority cell phenotype within each patch.
\textbf{C}, Visual comparison of KRONOS predictions against the ground truth phenotype map for a representative image from the DLBCL-1 dataset.
\textbf{D,E,F} KRONOS for region classification. 
\textbf{E}, ROC curves showing tumor vs. non-tumor classification using DAPI, cytokeratin, and androgen receptor (AR), a marker for prostate cancer cells, in a prostate cancer spatial proteomics dataset.
\textbf{F}, \textit{Left}: A spatial proteomics image showing DAPI, cytokeratin, and AR markers. \textit{Right}: KRONOS-based binary tumor prediction map overlaid on the same image.
\textbf{G,H,I} KRONOS for artifact detection. 
\textbf{H}, Class-level performance comparison of artifact detection using embeddings from DAPI. 
\textbf{I}, \textit{Left}: DAPI images from two spatial proteomics tissue sections. \textit{Right}: Corresponding KRONOS-based artifact detection prediction maps.
In \textbf{B} and \textbf{H}, each bar represents the average performance, with higher numbers indicating better performance, and error bars indicate the standard deviation across four folds. In \textbf{E}, solid lines represent mean performance, shaded areas indicate standard deviation across 4-folds, and the dotted line denotes random chance.}
\label{fig:patch_phenotyping}
\end{figure*}

\heading{Segmentation-free phenotyping}

Standard cell phenotyping currently involves cell segmentation\cite{stringer2021cellpose, greenwald2022whole, bai2021adjacent}, extraction of marker intensities for all cells, and clustering based on marker intensity profiles for phenotype assignment. While KRONOS enables strong phenotyping performance from pre-segmented cells, the standard approach remains sensitive to segmentation quality, imaging artifacts, and signal spillover from adjacent cells\cite{bai2021adjacent}.

As an alternative, we propose a new approach for segmentation-free phenotyping, where the inputs and the labels are defined at the patch level instead of the cell level, thereby removing the dependency on cell boundaries which can often be confounded by subjectivity or subpar performance due to the complexities of cell shapes and boundaries, including overlapping cells (\textbf{Figure \ref{fig:patch_phenotyping}A}). Specifically, by matching the patch size ($32 \times 32$ pixels or $16\mu m\times16\mu m$) to the average cell diameter, this approach can preserve a near single-cell resolution while bypassing segmentation-induced errors. We assigned patch labels to the majority cell class based on pixel coverage within annotated cell boundaries. The reliability of these labels depends on the proportion of patch pixels belonging to the dominant class. To evaluate performance across varying label quality, we tested this hypothesis by evaluating KRONOS on patches exceeding dominance thresholds of 20\%, 40\%, 60\%, 80\%, and 100\%. We compared our method to a standard baseline for cell classification based on mean marker expression\cite{jackson2020single, magness2024deep, shaban2024maps}, where each patch is represented by a feature vector containing the average expression of each marker. 

Segmentation-free, patch-based KRONOS cell phenotyping on DLBCL-1 dataset consistently outperformed the mean marker expression baseline across all thresholds (\textbf{Figure \ref{fig:patch_phenotyping}B,C}, \textbf{Supplementary Table \ref{tab:patch_phenotyping}}). The classification performance steadily improved as the ratio increased, as expected with lower label noise levels, e.g., balanced accuracy increasing from 0.5510 $\pm$ 0.0538 (threshold of 20\%) to 0.7183 $\pm$ 0.0759 (threshold of 100\%), compared to 0.4704 $\pm$ 0.0469 to 0.6885 $\pm$ 0.0759 for the marker expression baseline. This demonstrates a new property of KRONOS for efficient annotation of spatial proteomics data, and importantly, supports a new paradigm of patch-level spatial proteomics image analysis beyond conventional segmentation-based approaches. In addition, this approach introduces the potential for assigning more descriptive labels with scalable patch sizes, derived from a combination of multiple phenotypic labels, to embody richer biological context in each patch, building upon the concept of tissue microenvironments and cellular neighborhoods\cite{schurch2020coordinated}.

\begin{figure*}
\centering
\includegraphics[width=0.95\textwidth]{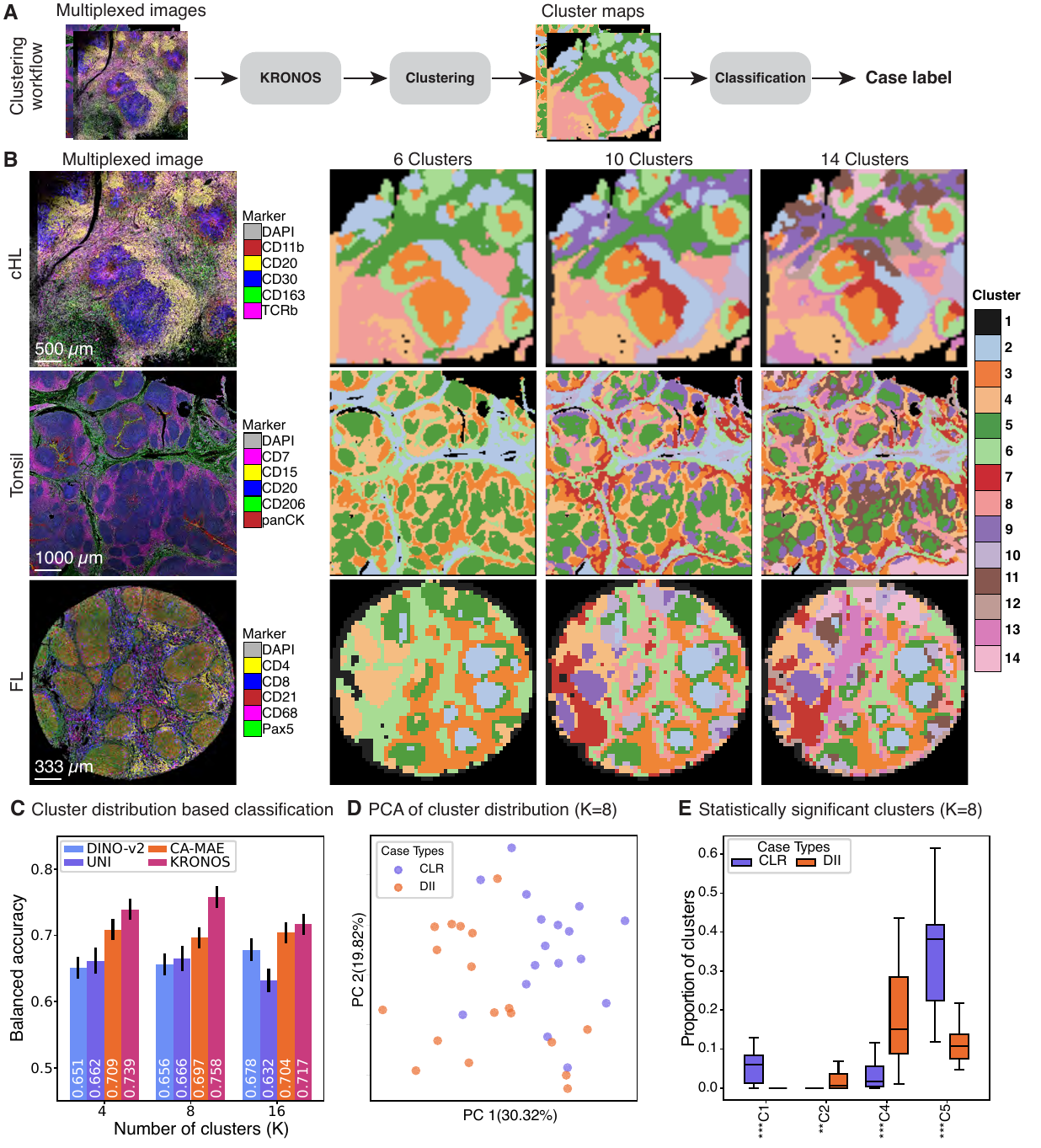}
\caption{\textbf{Unsupervised tissue phenotyping and classification with KRONOS.}
\textbf{A}, Overview of KRONOS for unsupervised clustering of multiplex images followed by k-nearest neighbor (kNN) classification from cluster distributions. 
\textbf{B}, Examples of clusters from classical Hodgkin lymphoma (cHL, top row), tonsil (middle row), and follicular lymphoma (FL, bottom row) multiplex images. were clustered into 6, 10, and 14 clusters with K-means clustering. Each color in the multiplex images encodes a distinct marker.
\textbf{C}, Balanced accuracy of KRONOS and baselines for predicting Crohn’s-like reaction (CLR) from diffuse inflammatory infiltration (DII) in patients with colorectal cancer (CRC).  Higher numbers indicate better performance, and error bars indicate the standard deviation across four folds.
\textbf{D}, Principal component (n=2 components) of patients from the CRC dataset\cite{schurch2020coordinated} were labeled by their clinical subtype: CLR vs. DII.
\textbf{E}, The four statistically significant clusters between CLR and DII patients in the CRC dataset.
Statistical significance was assessed with the Mann-Whitney U test.$^{\ast\ast}p\leq 0.01$,$^{\ast\ast\ast}p\leq 0.001$.
}
\label{fig:interpretability}
\end{figure*}

\heading{Patch-oriented tasks with KRONOS}

Shifting the basic biological entity from an individual cell to a larger image patch, akin to microenvironments, enables additional analyses for probing the underlying biology efficiently across entire spatial proteomics images using a "top-down" approach, rather than limited tissue niches with current analytical workflows. To this end, we further explore patch-oriented tasks at various image scales, including region classification, artifact detection, and unsupervised tissue phenotyping. 

\noindent\textbf{Region classification.} Spatial organization within the tissue microenvironment, such as immune infiltration zones, tumor margins, or tertiary lymphoid structures, encodes information relevant to patient prognosis, treatment response, and disease progression. Precise delineation of these regions enables localized analysis of spatial biomarkers and improves interpretability in downstream tasks. While manual annotation of these regions can be guided by distinct marker expression patterns (e.g., androgen receptor (AR) expression in prostate cancer), exhaustive annotation across large datasets remains labor and computationally intensive and subjective.

To evaluate KRONOS on region classification, we used a labeled prostate cancer dataset comprising 319 image regions ($256 \times 256$ pixels) from 112 tissue cores, annotated as tumor ($n=202$) or non-tumor ($n=117$). We trained a binary classifier using KRONOS embeddings derived from three channels: two tumor-associated markers (Cytokeratin and AR) and a nuclear stain, with expert-provided labels as ground truth (\textbf{Figure \ref{fig:patch_phenotyping}D}). KRONOS significantly outperforms the marker expression baseline, achieving an AUC of $0.91 \pm 0.03$ against $0.68 \pm 0.07$ (\textbf{Figure \ref{fig:patch_phenotyping}E}, \textbf{Supplementary Table \ref{tab:region_classification}}). Visual inspection confirmed that KRONOS accurately delineated tumor boundaries, with predicted regions closely matching areas of strong tumor marker expression (\textbf{Figure \ref{fig:patch_phenotyping}F}). 
The results demonstrate that patch-level KRONOS features can collectively identify spatial and biological signals at the region level beyond standard marker expressions based features.

\noindent\textbf{Artifact detection.} The complex data acquisition workflows associated with spatial proteomics, including tissue handling, staining, imaging, and pre-processing, can introduce a variety of artifacts that impact imaging quality and risk confounding downstream analysis. Common artifacts include tissue folding, blurring, antibody cross-reactivity, and pixel intensity saturation. While identifying artifacts is crucial for downstream analyses, artifact detection traditionally relied on cell segmentation and manual inspection by experts, which is time-consuming and prone to subjectivity, and is an emerging area for computational tools\cite{baker2024quality}.
KRONOS encoding capabilities bring opportunities for automatic artifact detection without the need for human intervention at the scale of the entire spatial proteomics image. 

To this end, we trained a 3-class classifier based on KRONOS embeddings on an annotated subset of the DLBCL-1 cohort (\textbf{Figure \ref{fig:patch_phenotyping}G}). The task involves classifying $256\times 256$ patches into artifact, background, or foreground regions. We focused specifically on identifying two prevalent artifacts using the nuclear stain channel (DAPI): blurring and signal saturation. The KRONOS classifier achieved a mean precision of $0.98\pm0.03$, outperforming the marker expression baseline ($0.74\pm0.02$) (\textbf{Figure \ref{fig:patch_phenotyping}H}, \textbf{Supplementary Table \ref{tab:artifact_classification}}). Visual inspection of the predicted artifact maps confirmed the accuracy of KRONOS segmented artifacts, foreground, and background, demonstrating its potential in detecting imaging inconsistencies across the entire tissue (\textbf{Figure \ref{fig:patch_phenotyping}I}).

\noindent\textbf{Unsupervised phenotyping with KRONOS clustering.} To evaluate whether KRONOS captures biologically meaningful spatial and tissue architectural structures, we performed unsupervised K-means clustering (K=6,10,14) of KRONOS features extracted from $64 \times 64$-pixel patches (\textbf{Figure \ref{fig:interpretability}A}). We performed clustering on three datasets: classical Hodgkin's Lymphoma (cHL), human tonsils, and follicular lymphoma (FL), each with different multiplexed marker panels (\textbf{Figure \ref{fig:interpretability}B}, \textbf{Online Methods}, section \textbf{Downstream datasets}). The resulting clusters highlight biologically meaningful tissue architecture and compartments. In cHL, tumor regions nested within immune-rich T cell zones and macrophage regions were observed in line with prior observations\cite{zhu2025celllens} (\textbf{Figure \ref{fig:interpretability}B}, top row). In tonsils, KRONOS-derived clusters delineated expected germinal centers, mantle zones, and interfollicular areas (\textbf{Figure \ref{fig:interpretability}B}, middle row). In FL, clusters captured the disrupted neoplastic follicular architecture and interfollicular areas (\textbf{Figure \ref{fig:interpretability}B}, bottom row). These results demonstrate that KRONOS representations encode biological spatial organization even in the absence of supervision and using efficient clustering algorithms such as k-means.

\noindent\textbf{Case-level stratification with KRONOS clustering.} To further evaluate the significance of KRONOS tissue clusters for TME analysis, we performed case-level binary classification based on the distribution of clusters from the colorectal cancer (CRC) dataset\cite{schurch2020coordinated}. The two patient groups within the dataset represent opposite extremes of immune TME architectures. The first group ($n=17$) exhibited a Crohn’s-like reaction (CLR), characterized by the de novo formation of numerous tertiary lymphoid structures (TLSs) at the tumor invasive front, whereas the other group ($n=18$) showed diffuse inflammatory infiltration (DII) and lacked TLSs. From these, we extracted KRONOS embeddings using a patch size of 256$\times$256 pixels with 50\% overlap, which were subsequently clustered using K-means on all cases. We computed the distribution of clusters for each case as the proportion of patches assigned to each cluster, resulting in the $K$-dimensional feature representing each case. 

Each test case was assigned the same label as that of the nearest case from the support set, as measured by Euclidean distance in the distribution space. Specifically, in each of 100 iterations, we randomly selected five cases from each group (a total of 10 cases) to form the support set. We benchmarked KRONOS against DINO-v2, UNI, and CA-MAE across different values of $K \in [4, 8, 16]$ (\textbf{Figure \ref{fig:interpretability}C} and \textbf{Supplementary Table \ref{tab:clustering_classification}}). KRONOS consistently outperformed all baselines. Specifically, at $K=8$, KRONOS reached its highest performance with 0.7584 $\pm$ 0.0162, outperforming 0.6564 $\pm$ 0.0159 (DINO-v2), 0.6665 $\pm$ 0.0191 (UNI), and 0.6952 $\pm$ 0.0157 (CA-MAE). 
Notably, the peak performance at $K=8$ aligns with the original findings\cite{schurch2020coordinated}, which identified nine distinct tissue neighborhoods.

Further visualization of the case-level cluster distributions using principal component analysis for $K=8$ showed separation between the CLR and DII groups (\textbf{Figure \ref{fig:interpretability}D}). Using the Mann–Whitney U test, we identified four clusters with statistically significant enrichment: two predominantly associated with CLR and two with DII (\textbf{Figure \ref{fig:interpretability}E}). Visual inspection of representative tissue cores (one for each of CLR and DII), previously selected by the original study authors, confirmed that regions belonging to the CLR-enriched clusters were localized within the CLR core, and similarly for the DII clusters (\textbf{Supplementary Figure \ref{fig:crc_clustering}}). 

\heading{Patient stratification with KRONOS}

\begin{figure*}
    \centering
    \includegraphics[width=1.00\textwidth]{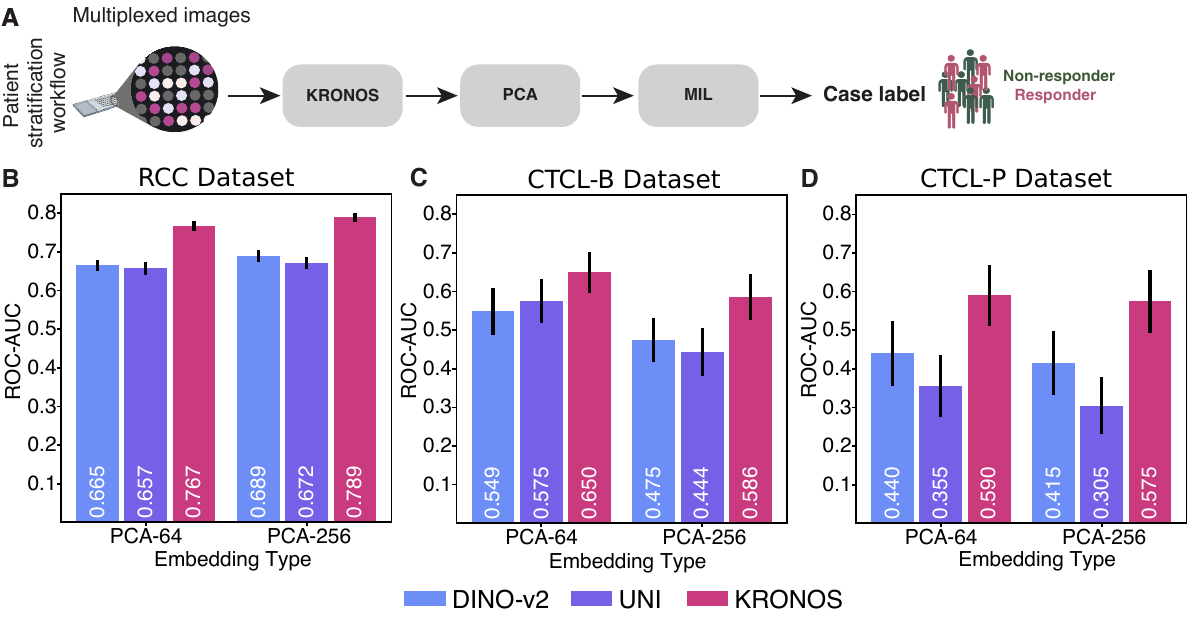}
    \caption{\textbf{Patient stratification with KRONOS.} 
    \textbf{A}, Overview of KRONOS for patient-level stratification using multiple instance learning (MIL) classification. 
    \textbf{B}, ROC-AUC classification of KRONOS against UNI and DINO-v2 for predicting response to immune checkpoint–based combination therapy in clear cell Renal Cell Carcinoma (ccRCC). The ccRCC cohort includes 27 patients and 108 tumor cores. 
    \textbf{C}, ROC-AUC classification of KRONOS for predicting response to antibody drug conjugate therapy in classical cutaneous T-cell lymphoma (CTCL-B). The CTCL-B cohort includes 28 patients and 56 tissue cores.
    \textbf{D}, ROC-AUC classification of KRONOS for predicting response to immunotherapy with pembrolizumab in CTCL patients (CTCL-P). The CTCL-P cohort includes 14 patients and 25 tumor cores.
  Higher numbers indicate better performance, and error bars indicate the 95\% confidence interval of results across 100 random train and test splits. 
    }
    \label{fig:downstream_analysis}
\end{figure*}

Spatial proteomics embodies a robust discovery tool for biomarkers that stratify between clinical outcomes, such as response to immunotherapeutics\cite{desouza2024multiplex}. We thus evaluated the potential of KRONOS embeddings for patient stratification of clinical response, specifically for classifying patients into responder and non-responder groups (\textbf{Figure \ref{fig:downstream_analysis}A}). Distinct from the previous approach of patient stratification with cluster distribution statistics, we use here a multiple instance learning (MIL)\cite{ilse2018attention,clam} approach where KRONOS patch features with each case are averaged to form a single case-level feature. This evaluation is performed on three multiplex imaging datasets: one clear cell renal cell carcinoma (ccRCC, \textbf{Figure \ref{fig:downstream_analysis}B}) dataset and two cutaneous T-cell lymphoma (CTCL) datasets, referred to as CTCL-B (brentuximab, \textbf{Figure \ref{fig:downstream_analysis}C}) and CTCL-P\cite{phillips2021immune} (pembrolizumab, \textbf{Figure \ref{fig:downstream_analysis}D}). All datasets contain more than 50 markers, resulting in high-dimensional embeddings. To address this, we use principal component analysis (PCA) to reduce the dimensions of the marker-specific embedding to either 64 or 256, before feeding the embeddings into a classifier. Each experiment uses 80\% of cases for training and 20\% for testing, repeated across 100 stratified random splits.

KRONOS consistently outperforms DINO-v2 and UNI across all datasets and PCA configurations tested here (\textbf{Figure \ref{fig:downstream_analysis}B,C,D} and \textbf{Supplementary Table \ref{tab:case_stratification}}). On the ccRCC dataset, KRONOS achieves the highest AUC of 0.7895 $\pm$ 0.0123 with PCA-256, outperforming DINO-v2 (0.6890 $\pm$ 0.0157) and UNI (0.6723 $\pm$ 0.0158). For the CTCL-B dataset, KRONOS reaches 0.6500 $\pm$ 0.0525 with PCA-64, compared to 0.5487 $\pm$ 0.0602 (DINO-v2) and 0.5750 $\pm$ 0.0576 (UNI). Similar trends are observed in CTCL-P, where KRONOS achieves an AUC of 0.5900 $\pm$ 0.0793 (PCA-64), significantly higher than both DINO-v2 (0.4400 $\pm$ 0.0850) and UNI (0.3550 $\pm$ 0.0804). These results demonstrate the superior discriminative capability of KRONOS embeddings for patient stratification tasks across diverse cancer types.

\begin{figure*}
    \centering
    \includegraphics[width=0.95\textwidth]{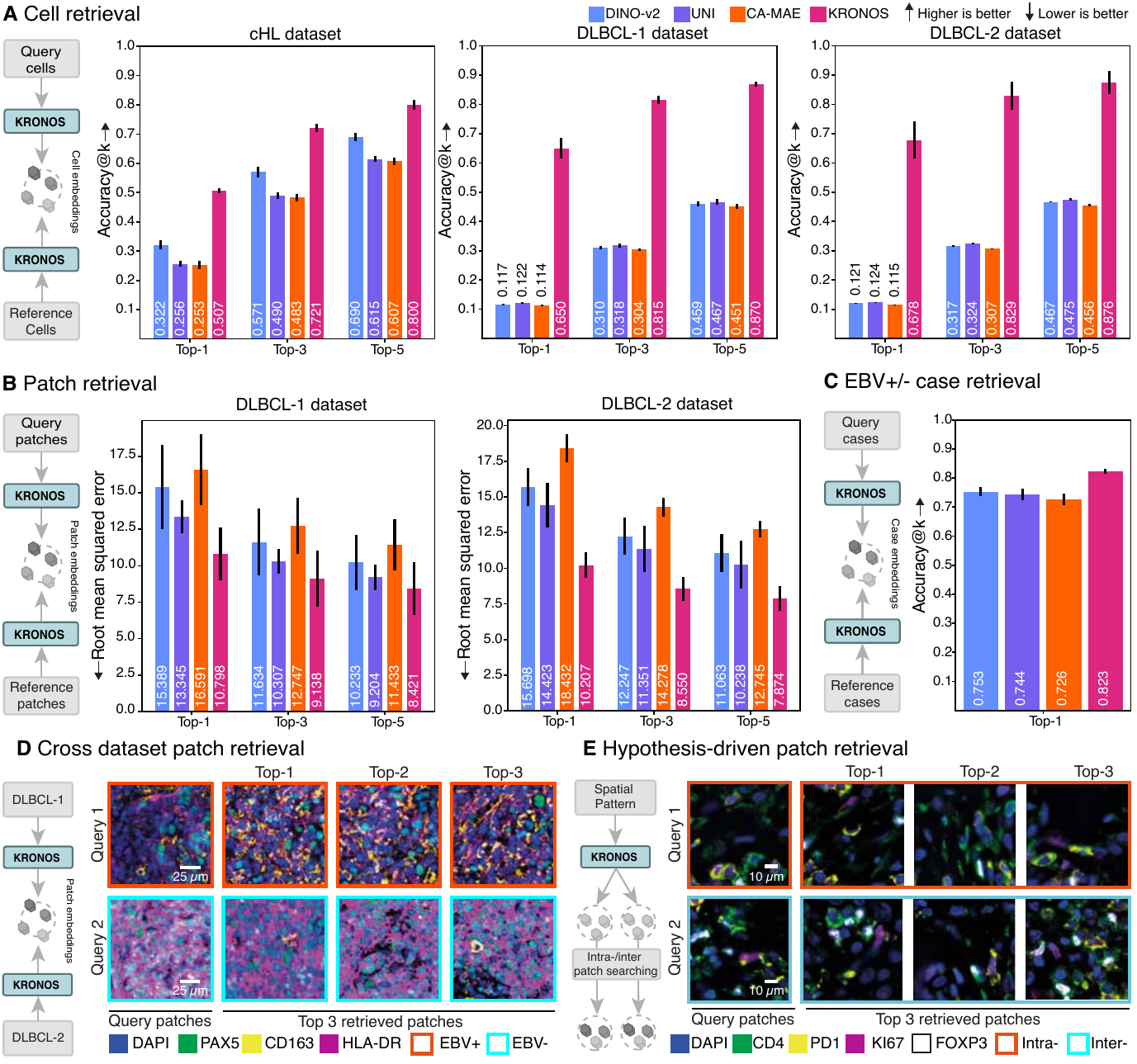}
    \caption{\textbf{Image retrieval using KRONOS embeddings.}
    \textbf{A}, \textit{Cell-level retrieval} measured with Accuracy@$k$ (higher the better) on the cHL, DLBCL-1, and DLBCL-2 cohorts. 
    Retrieval is deemed successful if at least one of the top-k retrieved images is of the same cell type as the query. Accuracy is computed as the proportion of the total query samples for which retrieval was successful.
    Error bars are calculated using four-fold cross-validation. 
    \textbf{B}, \textit{Patch-level retrieval} on DLBCL-1 and DLBCL-2 cohorts. Root mean squared error (lower is better) between the ground truth cell type distribution of the query and the retrieved patch is reported. When $k>1$, we use the best retrieved patch to compute the error. Error bars are calculated using four-fold cross-validation. 
    \textbf{C}, \textit{Case-level retrieval} of EBV-positive vs. EBV-negative samples on the combined DLBCL-1 and DLBCL-2 cohorts, using the nearest-neighbor search in the embedding space. Error bars are calculated using 100 random train and test splits. 
    \textbf{D}, \textit{Cross dataset patch retrieval} with DLBCL-1 and DLBCL-2 cohorts, where one cohort assumes the query set and the other assumes the support set. An example query patch of EBV-positive from DLBCL-2 and EBV-negative from DLBCL-1 is shown, along with the corresponding top-retrieved patches.
    \textbf{E}, \textit{Hypothesis-driven patch retrieval} targeting regions with tumor cells, Tregs, and PD-1$^{+}$ CD4$^{+}$ T cells, proposed as a spatial signature of anti–PD-1 response in CTCL. Intra-image retrieval uses queries and results from the same slide; inter-image retrieval pools across slides.
    }
    \label{fig:image_retrival}
\end{figure*}

\heading{KRONOS enables a spatial proteomics image reverse search engine}

Current spatial proteomics analysis faces significant challenges in comparing protein expression and cell patterns across different samples, experimental conditions, and datasets due to the lack of standardized computational tools for image-based pattern recognition and retrieval. Existing analysis pipelines primarily focus on quantitative measurements within individual experiments, but cannot efficiently scale to identify similar spatial protein distributions or cellular architectures between distinctive datasets, much less across large databases of spatial proteomics images. A reverse image search engine specifically designed for spatial proteomics would address this critical gap by enabling researchers to query complex protein localization patterns and discover functionally related protein networks, disease-associated spatial signatures, or evolutionarily conserved cellular organization patterns that reflect similar underlying biological processes across diverse experimental datasets and disease types.

Epstein-Barr Virus-positive (EBV-positive) and EBV-negative diffuse large B-cell lymphomas (DLBCL) exhibit fundamentally distinct tumor microenvironments, with EBV+ DLBCL characterized by increased immunosuppressive signaling, altered immune cell infiltration patterns, and a tolerogenic milieu that promotes immune evasion and dysfunction\cite{yeo2024sameslide,yeo2024epstein}. The stark differences in spatial protein expression patterns and cellular organization between EBV-positive and EBV-negative DLBCL tumor microenvironments present an ideal test case for validating the ability of KRONOS as an image reverse search engine to identify virus-associated spatial signatures.

\noindent\textbf{Cell retrieval.} We first apply KRONOS embeddings for cell retrieval using the top-k nearest neighbors (with $k$=1, 3, and 5) based on Euclidean distance in the embedding space to identify cells most similar to the query. Retrieved cells are then labeled as true or false positives based on whether their phenotype matches that of the query cell, and the retrieval performance Accuracy@$k$ is computed as the presence of at least one true positive within the top-k retrieved cells. For benchmarking, we use the training set of each cell phenotyping dataset (cHL, DLBCL-1, and DLBCL-2) as the reference set and the test set as the query set, following four-fold cross-validation. Across all three cell phenotyping datasets, KRONOS consistently outperforms baseline models (DINO-v2, UNI, and CA-MAE) (\textbf{Figure \ref{fig:image_retrival}A, Supplementary Table \ref{tab:cell_retrieval}}). On the cHL dataset, KRONOS achieves a top-1 accuracy of 0.5065$\pm$0.0086, significantly higher than the best baseline (DINO-v2 at 0.3217$\pm$0.0165), with even greater gains at top-3 (0.7214$\pm$0.0140) and top-5 (0.8004$\pm$0.0168) retrieval. On DLBCL-1 and DLBCL-2 datasets, KRONOS achieves top-1 accuracies of 0.6502$\pm$0.0340 and 0.6780$\pm$0.0637, respectively, more than five times higher than any competing method. 

\noindent\textbf{Patch retrieval.} Retrieval can also operate at the patch-level to query cellular patterns (or niches) that may reflect tumor progression, therapeutic resistance, and clinical outcomes\cite{teillaud2024tertiary,pelka2021spatially,schurch2020coordinated}.
Using the DLBCL-1 and DLBCL-2 datasets, we divided images into 256$\times$256 pixel patches, treating training patches as the reference set and test patches as the query set in a four-fold cross-validation setup. Patch similarity was measured using Euclidean distance in the embedding space, and retrieval quality was evaluated by comparing the cell phenotype distributions between query and retrieved patches using root mean squared error (RMSE); the lowest RMSE among the top-k retrieved patches was reported. KRONOS consistently outperformed baseline models (DINO-v2, UNI, and CA-MAE) across all $k$ settings and both datasets (\textbf{Figure \ref{fig:image_retrival}B, Supplementary Table \ref{tab:patch_retrieval}}). On DLBCL-1, KRONOS achieved RMSEs of 10.80$\pm$1.8082 (top-1), 9.14$\pm$1.9213 (top-3), and 8.42$\pm$1.8221 (top-5), substantially outperforming the best baseline scores. Similar gains were observed on DLBCL-2, where KRONOS achieved a top-1 RMSE of 10.21$\pm$0.9040, compared to 15.70$\pm$1.3068 (DINO-v2), 14.42$\pm$1.5871 (UNI), and 18.43$\pm$0.9948 (CA-MAE), with the top-5 RMSE improving to 7.87.

\noindent\textbf{Case retrieval.} We evaluated the ability of KRONOS embeddings to retrieve entire cases. To this end, we used 29 tissue cores (combining DLBCL-1 and DLBCL-2 datasets), each corresponding to a unique case labeled as either EBV-positive or EBV-negative. Case-level embeddings were computed by averaging the 256$\times$256 patch embeddings across all patches within a case. For benchmarking, we randomly selected three cases from each class as the reference set (a total of six cases) and treated the remaining cases as the query set, repeating this process 100 times. Similarity between cases was measured using Euclidean distance in the case embedding space, and only top-1 retrieval accuracy was used as the evaluation metric due to the binary nature of the labels. KRONOS outperformed all baselines across all evaluation metrics (\textbf{Figure \ref{fig:image_retrival}C, Supplementary Table \ref{tab:case_retrieval}}). It achieved the highest accuracy (0.8233$\pm$0.0097), compared to DINO-v2 (0.7529$\pm$0.0160), UNI (0.7436$\pm$0.0189), and CA-MAE (0.7262$\pm$0.0195).

\noindent\textbf{Cross-dataset patch retrieval.} We then investigated KRONOS's ability to perform patch- or tissue-niche-level search across datasets. Specifically, we selected a set of representative tissue immune microenvironment patches from the DLBCL-1 dataset and used them to query the DLBCL-2 dataset, and vice versa (\textbf{Figure~\ref{fig:image_retrival}D,E} and \textbf{Supplementary Figure~\ref{fig:retrieval_supp}}). The retrieval quality was evaluated based on three criteria between the query and the retrieved patches: 1) the EBV status of the tumor, 2) the proximity of aggregated marker expression profiles visualized in principal component space, and 3) the concordance of cell phenotype distributions.

Out of four query patches (two EBV-positive and two EBV-negative), three retrieved patches correctly matched the EBV status of the query patch (\textbf{Figure~\ref{fig:image_retrival}D} and \textbf{Supplementary Figure~\ref{fig:retrieval_supp}A}), showed high similarity in their aggregated marker expression profiles (\textbf{Figure~\ref{fig:image_retrival}E} and \textbf{Supplementary Figure~\ref{fig:retrieval_supp}B}), and cell phenotype distribution (\textbf{Supplementary Figure~\ref{fig:retrieval_supp}C,D}). For the one EBV-negative query patch, the top-3rd to top-5th retrieved patches originated from an EBV-positive case (\textbf{Supplementary Figure~\ref{fig:retrieval_supp}A}). Despite such a mismatch in EBV status, these three retrieved patches from the EBV-positive tumor visually resembled the query patch in terms of cell phenotype composition, and one showed similarity in aggregated marker expression profile. 

In parallel, we evaluated if KRONOS was able to retrieve spatial motifs within patches from patients with CTCL that responded to anti-PD-1 therapy \cite{phillips2021immune}. We successfully performed intra-tissue (query patch and search operate on the same image) and inter-tissue (query and search operate across patients) retrieval (\textbf{Figure~\ref{fig:image_retrival}E}) to identify patches containing the spatial motif consisting of a tumor cell, Treg, and PD-1 expressing CD4 T cell\cite{phillips2021immune}.

These results highlight the untapped potential of using foundation model embeddings, such as KRONOS, for scalable image search across spatial proteomics datasets. 

\heading{Robustness and design analysis of KRONOS}

\noindent\textbf{Ablation studies of KRONOS.} To evaluate the contribution of key architectural and design choices in KRONOS, we conducted a series of ablation experiments using the cHL and DLBCL-1 cohorts (\textbf{Supplementary Figure~\ref{fig:supp_ablation}}). Performance steadily improved with longer training, plateauing around 100,000–125,000 iterations (\textbf{Supplementary Figure~\ref{fig:supp_ablation}A}). Introducing a dedicated marker encoding (sinusoidal encoding for KRONOS) yielded the most pronounced gains, with up to a 37.4\% absolute increase in balanced accuracy on the cHL dataset (\textbf{Supplementary Figure~\ref{fig:supp_ablation}B}).  We further observed that using smaller input tokens (4$\times$4 pixels) improved accuracy compared to standard 16$\times$16 tokens, though this advantage could be matched by employing overlapping tokens with 50\% overlap (\textbf{Supplementary Figure~\ref{fig:supp_ablation}C,D}). Lastly, replacing the image-level (CLS token) embeddings with marker-specific embeddings, which is the number of markers longer than the CLS token, led to a substantial performance increase (\textbf{Supplementary Figure~\ref{fig:supp_ablation}E}). This prompted us to use of marker-specific embeddings for most of the analyses, unless computationally constrained. Collectively, these results emphasize the importance of marker-aware encoding and high-resolution spatial representation for robust cell phenotyping.

\noindent\textbf{Batch effect analysis.} Technical variability during sample preparation, such as differences in antibody staining temperature and duration, or variations in heat-induced epitope retrieval (HIER) buffers and timings, can introduce staining batch effects that obscure biological signals and compromise downstream analysis. To evaluate the robustness of KRONOS to such variation, we used a controlled dataset\cite{lee_2024_11410274} of 24 whole-slide multiplex images acquired using CODEX from consecutive tissue sections to assess the impact of biological variability. Experimental conditions were systematically varied across a combination of stain duration, antibody staining temperature, HIER duration, and HIER buffer types. Each image was then divided into $256\times256$ patches at 0.50~$\mu$m/px, from which KRONOS and mean marker expression embeddings were extracted. 

To quantify the batch effects, we computed the silhouette scores from the first two principal components of the combined set of two different experimental conditions, for every pair within 24 conditions. Based on aggregated scores across all 276 pairs, we observed that mean marker expression exhibited a larger batch effect as indicated by a higher silhouette score (0.0608~$\pm$0.0386) than KRONOS (0.0124$\pm$0.0307). This demonstrates that KRONOS effectively mitigates batch-associated variation, further supported by example visualizations in \textbf{Supplementary Figure~\ref{fig:batch_effect}}. This important aspect highlights future potential use cases of KRONOS embeddings to control for batch effects due to technical variations.

\noindent\textbf{KRONOS attention analysis.} To evaluate whether KRONOS learns biologically grounded representations, we further probe the self-attention mechanism within its ViT architecture, which forms the basis of representation learning. Specifically, we analyzed the spatial distribution of attention across markers. Each attention head highlights regions that contribute most to the learned embeddings, providing a way to probe the model’s internal focus. We computed marker-specific attention z-scores across multiple tissue types by aggregating attention values over cell regions defined by segmentation masks. For this analysis, we randomly sampled 1,000 cells per cell type from both the DLBCL-1 and DLBCL-2 datasets. Attention scores were aggregated per marker and visualized across cell types (\textbf{Supplementary Figure~\ref{fig:marker_attention}}). As expected, the model showed high attention for lineage-specific markers in the relevant cell types: PAX5 in B cells, CD31 in endothelial cells, FOXP3 in regulatory T cells, CD8 in cytotoxic T cells, CD15 in granulocytes and monocytes, and CD68 in macrophages. These results suggest that KRONOS encodes marker-level information in a biologically consistent manner, supporting its use for interpretable phenotyping and downstream tissue analysis.

\Heading{Discussion}

Our study demonstrates that large-scale self-supervised learning on spatial proteomics data yields general-purpose representations of cells, tissue regions, and entire tissue sections. KRONOS, our proposed foundation model, is trained on over 47 million image patches across 16 tissue types, 8 imaging technologies, and 175 protein markers.

KRONOS achieves strong performance across key benchmarks, including cell phenotyping, region classification, and artifact detection. It consistently outperforms general-purpose foundation models such as DINO-v2 (trained on natural images) and UNI (trained on histopathology), underscoring the need and applicability for spatial proteomics-specific models. Notably, KRONOS retains high accuracy even when trained with limited labeled data, highlighting its strong generalization and domain adaptation capabilities.

A major challenge in spatial proteomics is the limited availability of high-quality annotations due to small cohort sizes and heterogeneous study designs\cite{jackson2020single, schurch2020coordinated, amitay2023cellsighter}. KRONOS addresses this by providing label-efficient representations that support supervised learning with minimal labeled examples. Additionally, attention heatmaps generated by the model align with biologically meaningful cell types or tissue architecture, such as specific immune cell types or blood vasculature, demonstrating its interpretability and utility in exploratory analysis.

Spatial proteomics studies are characterized by various technologies, antibody panels, imaging platforms, data artifacts, tissue types, and study sites. KRONOS overcomes these challenges by simultaneously collapsing batch effects, supporting segmentation-free patch-based phenotyping, and enabling complex downstream analyses. These capabilities span from cell type to region and patient stratification, to image search and retrieval, regardless of platform, panel, or tissue type. We envision the resulting unified pipeline will be transformative in scaling spatial proteomics across panels and sites for discovery and clinical applications.

While our training dataset is diverse in tissue types and fluorescence-based imaging platforms, it does not currently include ion-based modalities such as imaging mass cytometry (IMC\cite{giesen2014imc}) or multiplexed ion beam imaging (MIBI\cite{angelo2014mibi}). These modalities differ significantly in both marker resolution and imaging outputs, and may require architectural adjustments or modality-specific adaptation strategies for future iterations of KRONOS.

KRONOS sets the foundation for scalable, annotation-efficient analysis in spatial proteomics. We further demonstrate the paradigm of segmentation-free, patch-level analysis of spatial proteomics data akin to histopathology approaches. Its pretrained backbone can facilitate future downstream tasks such as spatial biomarker discovery, spatial trajectory modeling, multimodal integration with transcriptomics or clinical data, and virtual tissue work. We envision future efforts will focus on expanding its modality coverage and incorporating temporal or longitudinal data to support dynamic tissue modeling.

\newpage
\Heading{Online Methods}

\heading{KRONOS pretraining and modeling}

\hheading{Data collection and cleaning}

KRONOS was pretrained on SPM-47M, a collection of diverse multiplex spatial proteomics samples sourced from public (56\%) and proprietary (44\%) cohorts across eight fluorescence-based imaging technologies and 16 tissue types. The cohorts differ significantly from each other in marker panels, tissue types, and disease contexts. The number of protein markers per cohort ranged from 3 to 52, with a total of 175 unique markers (\textbf{Figure \ref{fig:overview}B, Supplementary Table \ref{tab:pretrain_dataset}, \ref{tab:marker_list}}).

To mitigate artifacts from slide preparation and scanning, such as tissue folds, blurring, and intensity saturation, we manually reviewed each marker and masked visible defects (\textbf{Figure \ref{fig:overview}A}). Staining quality was assessed by comparing each marker to reference data from antibody vendors or the Human Protein Atlas. Markers with absent or incorrect staining patterns were excluded. Each image was then divided into patches (or tiles), denoted as $\largex\in\mathbb{R}^{256\times 256 \times M}$ where $M$ is the number of markers. To increase diversity, images were patched with 50\% overlap. This resulted in KRONOS pretraining dataset, SPM-47M, a collection of 47 million single-marker patches extracted from 3.67 million multiplexed patches (\textbf{Supplementary Table \ref{tab:pretrain_dataset}}).

\hheading{KRONOS modeling}

KRONOS is a Vision Transformer (ViT\cite{vaswani2017attention,vit})-based foundation model (ViT-S/16) specialized in processing multiplexed spatial proteomics images. It produces low-dimensional representations of input patches. KRONOS pretraining builds on DINO-v2, a self-supervised learning (SSL) method widely employed for training foundation models for pathology\cite{uni, phikon, hibou,  virchow, virchow2, giga-path}. KRONOS takes as input a multiplexed image $\largex \in \mathbb{R}^{H \times W \times M}$, where $H$ and $W$ denote the spatial dimensions and $M$ the number of markers. It is then tessellated into $N = (H/P) \times (W/P)$ non-overlapping square tokens of size $P \times P$, where we set $P = 16$. This results in a set of $N$ spatial tokens for each of the $M$ channels, denoted ${\x_i^j}$, where $\x_i^j \in \mathbb{R}^{P \times P}$ represents the $i$-th token for the $j$-th marker.

As samples contain a variable number of markers, the traditional RGB-channel processing widely used in vision foundation models\cite{dinov2, mae, ibot, i-jepa, uni, phikon, hibou,  virchow, virchow2, giga-path} cannot be applied. In addition, at inference time, the model should be able to integrate new markers, unseen during pretraining. To address these challenges, KRONOS introduces a shared convolutional filter and marker-specific encoding. First, the initial convolutional layer of the ViT is modified so that the same convolutional weights are shared across all markers. Specifically, the token embedding $\bar{\x}_i^j\in\mathbb{R}^D$ is provided as
\begin{equation}
    \bar{\x}_i^j = f_{\text{embed}}(\x_i^j)\in\mathbb{R}^D\quad \forall i, j,
\end{equation}
where $f_{\text{embed}}(\cdot)$ represents the shared convolutional embedding layer.
This allows the model to build marker-specific tokens, rather than processing all markers jointly. This enables encoding to any number, type, and order of markers, including those unseen during the pretraining stage. 

Next, we introduce non-learnable marker encodings $\{\m_j\}_{j=1}^M$, such that each token can be uniquely identified by its marker and spatial location. Inspired by the sinusoidal positional encoding employed in Transformers\cite{vaswani2017attention, chen2021empirical}, we map each marker to a corresponding sinusoidal encoding. Specifically, the $j^{\text{th}}$ marker encoding is defined as 
\begin{equation}
\m_{j, d} = \sin(j/10000^{2d/D})\quad \m_{j, d + D/2} = \cos(j/{10000^{2d/D}}),
\end{equation}
where $d$ ranges from $0$ to $\frac{D}{2} - 1$. This formulation ensures that $\m_j\neq \m_{j'}$ for $j\neq j'$, and thus enables effective marker differentiation. Moreover, since the marker encodings are not learned, new unseen markers can seamlessly be incorporated at the inference time by mapping them to an unused marker index in the sinusoidal encoding.

Lastly, we add a learnable positional encoding $\phi_i \in \mathbb{R}^{D}$ to encode 2D location $i$ and distinguish the spatial arrangement of tokens. The final token embedding $\z_i^j$ for marker $j$ of token $i$ is given by, 
\begin{equation*}
    \z_i^j = \bar{\x}_i^j + \m_j + \phi_i.
\end{equation*}
Each token becomes contextualized by both its spatial location (with $\phi_i$) and its marker (with $\m_j$). 

The set of embeddings $\{\z_i^j\}_{i,j=1}^{N,M}$ and a single learnable class ($\operatorname{CLS}$) token, $\z_{\text{CLS}}\in\mathbb{R}^D$, are subsequently fed into a series of 12 Transformer blocks for processing via self-attention. As a result, KRONOS produces contextualized embeddings $\{\bar{\z}_i^j\}_{i,j=1}^{N,M}$, with $\bar{\z}_i^j\in\mathbb{R}^D$, and $\bar{\z}_{\text{CLS}}\in\mathbb{R}^D$. The resulting embeddings are used in three ways:
\begin{enumerate}
    \item \textbf{Image-level features.} The $\operatorname{CLS}$ token $\bar{\z}_{\text{CLS}}\in\mathbb{R}^D$ is used as a single patch-level feature that summarizes the input multiplexed image $\largex$.
    \item \textbf{Marker-specific features.} For each marker $j \in {1, \dots, M}$, we compute a spatially aggregated feature $\bar{\z}^j$ by averaging over all spatial locations: $\bar{\z}^j= \frac{1}{N}\sum_{i=1}^N \bar{\z}_i^j$. This results in a set of $M$ marker-specific embeddings ${\bar{\z}^j}$ that summarize spatial information in $\largex$ for each marker. In downstream tasks, we concatenate the $M$ vectors into a single $MD$-dimensional image-level embedding.
    \item \textbf{Token-specific features.} The set of tokens concatenated along the marker dimension $\{[\bar{\z}^1_i,\ldots, \bar{\z}^M_i]\}_{i=1}^{N}$, where $[\cdot]$ denotes concatenation, can be used to represent location-specific information across different markers.
\end{enumerate}

At inference time, KRONOS can encode any multiplex image with an arbitrarily large number of markers. Using the positional encoding interpolation, the input image size can also be adjusted based on the downstream application to encode a cell or a tissue region. 

\hheading{KRONOS training}

\noindent\textbf{Pretraining batch preparation.} Each batch is built to include three markers per image: one nuclear stain marker (e.g., DAPI) and two randomly selected protein markers. Including a nuclear stain ensures that the model can localize cells and distinguish empty regions from cells that do not express the marker. Limiting the input to three channels balances representation diversity with memory constraints, as additional channels increase GPU memory usage.

To increase training diversity, we applied data augmentation by randomly cropping a \(224 \times 224\) region from each \(256 \times 256\) input image, combined with random horizontal and vertical flips, and random rotations of \(0^\circ\), \(90^\circ\), \(180^\circ\), or \(270^\circ\). In addition, each marker was normalized using the mean expression values and their standard deviations computed across the entire pretraining dataset.

SPM-47M spans a wide range of tissue types and imaging platforms, leading to significant imbalances across tissue-technology combinations. For example, lymph node patches imaged with CellDive were overrepresented compared to skin patches from IBEX (\textbf{Supplementary Table \ref{tab:pretrain_dataset}}). To prevent overfitting, we designed a two-stage stratified sampling strategy. First, SPM-47M was partitioned into 23 subsets, each containing one or a few tissue-technology pairs grouped by sample availability. During training, patches were sampled hierarchically: a subset was randomly selected, then a patch was drawn uniformly from within that subset. This approach exposed the model to a more balanced distribution of tissue-technology pairs and mitigated bias toward overrepresented sources.

\noindent\textbf{Pretraining objective.} KRONOS employs a self-supervised student–teacher framework based on DINO-v2\cite{dinov2}. DINO-v2 combines two objectives: self-distillation and masked image modeling (MIM), with both terms operating on two independently augmented views of the same input multiplexed image. For the self-distillation objective, the model minimizes the cross-entropy loss between the $\operatorname{CLS}$ token predictions of the student and teacher networks. Each network processes a distinct augmented view, and the teacher is updated as an exponential moving average (EMA) of the student to ensure stable training. In parallel, the MIM objective follows an online tokenizer strategy. Specific spatial regions are masked across all marker channels in the student input, and the model is trained to reconstruct the corresponding patch tokens from the teacher. The cross-entropy loss is computed between the masked token outputs of the student and the corresponding patch token predictions of the teacher, which serves as a dynamic, context-aware tokenizer. KRONOS is the teacher network.

\noindent\textbf{Pretraining parameters.} We trained KRONOS for 125,000 iterations with a batch size of 1,024, following the hyperparameter configurations recommended in the previous study\cite{uni}. All pretraining hyperparameters are detailed in \textbf{Supplementary Table \ref{tab:kronos_config}}.

\heading{KRONOS evaluation}

\hheading{Downstream tasks}

\noindent \textbf{Cell phenotyping.} To evaluate KRONOS for cell phenotyping, we extract a \(64 \times 64\) pixel patch from selected markers, centered at the cell's nucleus using a cell mask obtained with Mesmer\cite{tissuenet}. To ensure that each patch only contains information from a single cell, we zero-masked all pixels outside of the target cell mask. Each patch is then passed through KRONOS to extract marker features, which were then used for downstream multiclass classification with linear probing. Linear probing models were trained using the L-BFGS solver with \(\ell_2\)-regularization using the cuML library. The optimal value of the regularization hyperparameter (\(C\)) was determined using Bayesian optimization across a range of \(10^{-10}\) to \(10^5\) over 25 trials. The model achieving the highest \(F_1\) score in the validation set was selected as the optimal model and subsequently evaluated in the test set.

\noindent \textbf{Few-shot cell phenotyping.} To evaluate the label efficiency of KRONOS, we perform linear probing with limited labeled samples. Specifically, we constructed four training subsets containing 100, 200, 500, and 1000 randomly selected cells per phenotype. Each training subset was then used to train a linear probing model and evaluated on the full test set.

In addition, we evaluated KRONOS for human-guided cell phenotyping by simulating scenarios where only a small number of labeled cells are available within the image.. For each image in the DLBCL-1 dataset, we first excluded cell classes with fewer than 50 cells to ensure sufficient data for both training and validation. From the remaining classes, we randomly sampled 20 cells per class to form the initial training set (20-shot), and used all other cells from the same image as the validation set. The 20-shot training set was then subsampled to generate 10-shot, 5-shot, and 1-shot subsets. To account for sampling variability, we repeated the procedure 100 times for each image and each shot configuration. For each trial, a separate logistic regression model was trained per image using KRONOS features as input. Unlike previous few-shot experiments, we did not use cross-validation and instead trained one model per image. The regularization strength was fixed at 1, after testing values of 1, 0.01, 0.001, and 0.0001 to avoid the computational overhead of Bayesian optimization. Results were aggregated across all images and trials.

\noindent \textbf{Zero-shot evaluation.} To evaluate KRONOS in a zero-shot setting, we assessed whether the learned clusters in the cell embedding space align with their corresponding cell types. To this end, we perform K-means clustering with the number of clusters set to the number of annotated cell types (16 classes for cHL and 9 classes for DLBCL-1 and DLBCL-2). We measure the adjusted rand index (ARI) of the resulting clusters, with the higher value indicating that the learned clusters align better with the cell type labels.

\noindent \textbf{Patch-level segmentation-free phenotyping.} To assess whether KRONOS patch-level embeddings can capture dominant cell phenotypes without relying on explicit cell segmentation, we designed a segmentation-free classification task. The objective was to assign a single phenotype label to each image patch based on the most represented cell type within it. Specifically, overlapping patches of size \(32 \times 32\) pixels were extracted from the DLBCL-1 dataset using a stride of \(16 \times 16\) pixels. For each patch, the label was determined by first mapping each pixel to the label of its associated cell (using ground-truth segmentation masks), followed by majority voting across all pixels within the patch.

To ensure consistency with the cell phenotyping task, patches were split according to the same four-fold cross-validation scheme. KRONOS embeddings were computed for each patch and used to train a linear classifier. To evaluate the effect of label noise introduced by heterogeneous patches, we reported model performance across subsets of patches filtered by minimum majority-class pixel ratio thresholds (e.g., 20\%, 40\%). The higher threshold implies that the label noise levels are low, as a higher ratio of pixels belongs to the majority class.

\noindent \textbf{Region classification and artifact detection.} For each spatial position, we concatenate the spatial tokens of all markers at each location, to form the token-specific embedding. The use of token-specific embeddings enables spatially denser predictions, enabling high-resolution inference across entire images. We also performed morphological operations (opening and closing) to fill holes in the final prediction map. 

\noindent \textbf{Unsupervised clustering of tissue.} For unsupervised clustering, each image was divided into $64\times64$ pixel patches, from which we extracted the image-level features (i.e., CLS token). The use of the image-level $D$-dimensional features instead of the $MD$-dimensional marker features was chosen to reduce computational requirement. We then applied K-means clustering to the patch features within each image. The resulting cluster labels were mapped to the original images to visualize how the clustering patterns align with the underlying marker expression patterns.

For case startification for using case cluster distributoin, we use a patch size of $256\times256$ pixel with 50\% overlap and used marker-specific featues for K-means clustering. k-NN method is then used to classify cases test sets by finding their nearest neighbour in the train set. Cases in train and test set were assigned based on strafied random sampling, in 100 trails.



\noindent \textbf{Batch effect analysis.} To analyze the batch effect due to technical variations, evaluation was performed on every possible pair of 24 serial sections\cite{lee_2024_11410274} processed with different protocols (total of 276=$24 \choose 2$ pairs). For evaluation with silouhette index and adjusted rand index (ARI), KRONOS patch features from two sections were pooled together. Silhouhette index was computed on patch embeddings projected onto the first two principal components. ARI was computed on patch embeddings without further dimensionality reduction.

\noindent \textbf{Model interpretability analysis.} To analyze model interpretability, we extracted the attention scores of the CLS token from the final Transformer blocks for all marker tokens for the first attention head. The overall attention score of a given marker is computed by summing the attention scores of all tokens associated with that marker. For improved interpretability, the attention scores were standardized into z-scores, and boxplots were generated to visualize the distribution of z-scores across cell types per marker. For visualization, we selected 256$\times$256-pixel patches that contain a diverse mix of cell types. We then displayed protein marker images alongside token-level attention scores for the corresponding marker.

\hheading{Downstream datasets}


\noindent\textbf{Classical Hodgkin Lymphoma (cHL):} The cHL dataset\cite{shaban2023maps,shaban2024maps} comprises a single large CODEX-imaged field of view (approximately \(8,000 \times 8,000\) pixels at 0.37$\mu$m/px pixel resolution) from one patient. cHL includes 134,552 segmented cells annotated into 16 cell phenotypes. A total of 43 markers were acquired, of which 18 were used in this study. For each cell type, over 8,000 cells were annotated. The cHL image was divided into four equal sized quardrants to generate four folds for cell phenotyping task. This dataset is publicly available and serves as a benchmark for cell phenotyping under various evaluation scenarios. Marker panels are listed in \textbf{Supplementary Table~\ref{tab:phenotyping_markers}}.


\noindent\textbf{Diffuse Large B-Cell Lymphoma 1 (DLBCL-1):} The DLBCL-1 cohort comprises 60 large multiplexed CODEX-imaged tissue cores (on average \(4,166 \times 3,681\)-pixel at 0.50$\mu$m/px) collected from patients diagnosed with Diffuse Large B-Cell Lymphoma. A total of 27 markers were acquired, among which 12 were used for cell type annotation in 17 cores, resulting in 361,341 cells annotated into nine cell phenotypes. In addition, we annotated 199 regions of size \(256 \times 256\) from 45 cores for artifact detection (n=87 as foreground, n=81 as artifact, n=31 as background). Finally, we annotated 14 cores for Epstein-Barr virus (EBV) status prediction (n=6 EBV-positive and n=8 EBV-negative). The dataset is divided into four folds where each core is only part of one fold. Marker panels are listed in \textbf{Supplementary Table~\ref{tab:phenotyping_markers}}.


\noindent\textbf{Diffuse Large B-Cell Lymphoma 2 (DLBCL-2):} The DLBCL-2 cohort comprises 25 large CODEX-imaged tissue cores (on average 6161$\times$6163-pixel at 0.50$\mu$m/px by core). A total of 27 markers were acquired, among which 11 were used for cell type annotation in 21 cores, resulting in 31,098,443 cells annotated into nine cell phenotypes. In addition, 16 cores were annotated as EBV-positive (n=11) and EBV-negative (n=5). The dataset is divided into four folds where each core is only part of one fold. Marker panels are listed in \textbf{Supplementary Table~\ref{tab:phenotyping_markers}}.


\noindent\textbf{Prostate Cancer:} The prostate cancer dataset includes 319 image regions ($256 \times 256$ pixels), labeled as tumor (n=202) or non-tumor (n=117), annotated from 112 cores across 4 TMAs. For feature extraction, we used only three markers (DAPI, Androgen Receptor, and Cytokeratin) and employed spatial features instead of marker features for linear probing. It was acquired using CODEX at 0.50$\mu$m/px with 48 markers, among which three are used for tumor prediction (DAPI, androgen receptor, cytokeratin).

\noindent\textbf{Tonsil:} The human tonsil dataset consists of a full tissue section stained with a 48-plex antibody panel and imaged using CODEX (0.37 $\mu$m/px). We use this dataset for unsupervised clustering.

\noindent\textbf{Follicular Lymphoma (FL):} The FL dataset consists of a representative TMA core stained with a distinct 48-plex antibody panel and imaged using CODEX (0.37 $\mu$m/px). We use this dataset for unsupervised clustering.


\noindent\textbf{Batch effect analysis dataset:} From a single tissue block, 24 serial tissue sections were cut and imaged with CODEX under varying conditions\cite{lee_2024_11410274}. Experimental conditions were systematically varied across stain duration (1 hour vs. overnight), stain temperature (4$^\circ$C, 25$^\circ$C, 37$^\circ$C), HIER duration (10, 20, 40 minutes), and HIER buffer type (Citraconic Anhydride, pH 7.4; Tris/EDTA, pH 9.0). 



\noindent\textbf{CRC\cite{schurch2020coordinated}:} The CRC dataset includes CODEX imaging of 56 proteins (at 0.375$\mu$m/px) across 140 tissue regions collected from the tumor invasive front of 35 advanced-stage colorectal cancer (CRC) patients. These patients were selected from an initial cohort of 715 CRC cases, excluding those with low-stage disease (pTNM 0–2), prior chemotherapy, insufficient tissue material, or low immune infiltration. The final cohort consists of 17 patients with Crohn’s-like reaction (CLR), marked by a high density of tertiary lymphoid structures (TLS), and 18 patients with diffuse inflammatory infiltration (DII) and no TLS. This dataset is publicly available\cite{schurch2020coordinated}.

\noindent\textbf{Cutaneous T-Cell Lymphoma (CTCL-B):}  This dataset comprises 56 diagnostic pretreatment tissue cores collected from 28 patients with classical cutaneous T-cell lymphoma, all of whom received antibody-drug conjugate therapy (brentuximab vedotin). In addition, 28 matched on-therapy cores were available for 14 of these patients, resulting in a total of 84 tissue samples. Among the patients, 20 were classified as responders (R) and 8 as non-responders (NR). This dataset was used for patient stratification based on treatment response.

\noindent\textbf{Clear Cell Renal Cell Carcinoma (ccRCC):} This dataset includes 108 tumor tissue cores obtained from 27 patients with histologically confirmed ccRCC treated with immune checkpoint–based combination therapy. A total of 18 cores from 4 patients received cabozantinib plus nivolumab (CABO-NIVO), all categorized as non-responders (NR). The remaining 90 cores were collected from 23 patients treated with ipilimumab plus nivolumab (IPI-NIVO), including 41 cores from 10 responders (R) and 49 cores from 13 non-responders (NR). This dataset was used for patient-level treatment response prediction.

\noindent\textbf{Cutaneous T-Cell Lymphoma (CTCL-P):} This dataset includes 25 tumor tissue cores obtained from 14 patients diagnosed with cutaneous T-cell lymphoma (CTCL-P), all of whom were treated with the immune checkpoint inhibitor pembrolizumab as part of a clinical trial. Of the 14 patients, 7 were classified as responders (R) and 7 as non-responders (NR), resulting in 11 response cores and 14 non-response cores. Tumor cores were imaged with CODEX using 51 markers.



\hheading{Evaluation baselines}

We compare KRONOS against several baselines: DINO-v2\cite{dinov2}, UNI\cite{uni}, CA-MAE\cite{ca-mae}, and the mean marker expression.

\noindent \textbf{DINO-v2\cite{dinov2}:} We use the Vision Transformer-Large (ViT-L/14) model from DINO-v2, pretrained on the LVD-142M dataset of natural images. To adapt this RGB-based model to multiplexed data, each marker channel from a patch is individually replicated across the three RGB channels and passed through the model. For each marker, we extract the corresponding class token (CLS), and concatenate the CLS embeddings across all $M$ markers, resulting in a feature vector of size $1024 \times M$. Because DINO-v2 operates with a token size of $14\times14$ pixels and expects image dimensions that are multiples of 14, we apply center cropping to each marker image to satisfy this constraint. Additionally, as a preprocessing step, we normalize each marker channel using marker-specific means and standard deviations computed from the SPM-47M dataset.

\noindent \textbf{UNI\cite{uni}:} We use the Vision Transformer-Large (ViT-L/16) model from UNI, pretrained with DINO-v2 on the Mass100K histology dataset. As with DINO-v2, each marker channel is converted to RGB format by replicating it across all three channels. These images are passed independently through the model to extract the corresponding class token (CLS) for each marker. The resulting CLS embeddings are concatenated to form a feature vector of size $1024 \times M$, where $M$ is the number of markers. All marker channels are normalized using marker-specific mean and standard deviation values computed from the SPM-47M dataset, consistent with our preprocessing pipeline.

\noindent \textbf{CA-MAE\cite{ca-mae}:} We use the channel-agnostic masked autoencoder (CA-MAE), a ViT-based model pretrained on fluorescence-based cell profiling datasets (RxRx3\cite{fay2023rxrx3} and JUMP-CP\cite{chandrasekaran2023jump}). Unlike standard ViTs trained on RGB images, CA-MAE is natively designed to accept inputs with arbitrary channel counts and orders. To extract features from a multiplexed patch, we stack the marker channels as-is and pass the full tensor through the encoder. Each marker channel is normalized using marker-specific statistics computed on the SPM-47M dataset, consistent with our preprocessing pipeline. We use the marker-wise embeddings returned by CA-MAE of size $384 \times M$, where $M$ is the number of markers.

\noindent \textbf{Mean marker expression:} A widely used approach in multiplexed image analysis is to summarize protein expression by computing the mean intensity of each marker channel within a segmented cell~\cite{jackson2020single, magness2024deep, shaban2024maps}. This per-marker mean forms a vector of size $1 \times M$, where $M$ is the number of markers, and serves as a simple, non-spatial representation of cellular phenotype. We extend this principle beyond individual cells to full image patches, computing the mean intensity across all pixels in a patch for each marker channel. This yields a comparable $1 \times M$ feature vector that captures average marker expression at the patch level, independent of cell segmentation. For region classification and artifact detection, we use raw marker intensities as feature due to the use of limited marker for these two task: 3 markers for region classification and one nuclear marker for artifact detection.

\hheading{Statistical analysis}

For cell phenotyping, region classification, artifact detection tasks, cell retrieval, and patch retrieval, we employed a 4-fold cross-validation setup to calculate performance metrics, reporting the average performance along with the standard deviation across all folds. For a given experiment, one fold was used as the test set, while the remaining three folds were split into training (80\%) and validation (20\%) sets.
We follow the evaluation setup commonly used in representation learning benchmarking\cite{dinov2,uni}. Specifically, we perform linear probing (logistic regression) between frozen image embeddings and downstream labels. 

For case retrieval, case classification, and patient stratification tasks, we randomly assign cases into train and test sets in 100 trials and report the mean performance with 95\% confidence interval.
\hheading{Evaluation metrics}

Multiclass downstream tasks are evaluated using balanced accuracy, macro \(F_1\)-score, macro average precision, and the one-vs-one macro area under the receiver operating characteristic curve (AUROC). Binary downstream tasks are evaluated using balanced accuracy, \(F_1\)-score, and AUROC.

\heading{Computing hardware and software}

All experiments and analyses in this study (unless otherwise specified) were conducted using Python (v3.9.0) and PyTorch (v2.0.0, CUDA 12.0) \url{https://pytorch.org/}, with full reproducibility enabled through the use of open-source libraries as described below. To train KRONOS, we modified the original DINO-v2 algorithm implemented by Facebook Research and available at \url{https://github.com/facebookresearch/dinov2}. Pretraining was performed on 8$\times$80GB NVIDIA A100 GPUs configured for multi-GPU training using Fully Sharded Data Parallel (FSDP), while all downstream experiments were conducted on a single 24GB NVIDIA 3090 GPU. Multiplex image processing was performed using the \texttt{tifffile} package. For training downstream models, including logistic regression, we employed \texttt{cuML} (v24.12.00) and \texttt{scikit-learn} (v1.5.2). Several additional Python libraries were used throughout our pipeline: \texttt{pandas} (v2.2.2) for data manipulation and analysis; \texttt{h5py} (v3.12.1) for reading and writing HDF5 files used to store high-dimensional embeddings; \texttt{scipy}(v1.14.1) for scientific computing tasks, including matrix operations and clustering; and \texttt{optuna}(v4.1.0) for hyperparameter optimization. For data visualization, we used \texttt{matplotlib}(v3.10.0) and \texttt{seaborn}(v0.13.2) to generate analytical plots and statistical figures. Final figures were composed and refined using BioRender and Adobe Illustrator.

\heading{Data and code availability}

The code and model weights for loading KRONOS are available for academic research purposes at: \url{https://github.com/mahmoodlab/kronos}. Public pretraining datasets can be accessed from the following sources: the spleen and thymus dataset generated using CODEX is available through the HuMAP portal (\url{https://portal.hubmapconsortium.org/}); the ImmunoAtlas dataset can be downloaded from the ImmunoAtlas portal (\url{https://immunoatlas.org/}); the IBEX dataset is available on Zenodo (\url{https://zenodo.org/records/5244551}); and the CellDive dataset can be accessed via the Image Data Resource (IDR) portal (\url{https://idr.openmicroscopy.org/}). For public evaluation datasets, the cHL dataset is available on Zenodo (\url{https://zenodo.org/records/10067010}), the CRC dataset can be downloaded from The Cancer Imaging Archive (\url{https://doi.org/10.7937/tcia.2020.fqn0-0326}), and the CTCL-pembrolizumab dataset is available via the ImmunoAtlas portal (\url{https://immunoatlas.org/}). The batch effect analysis dataset is available on Zenodo (\url{https://zenodo.org/records/11410274}). Unpublished pretraining and evaluation datasets used in this study are not publicly available due to institutional restrictions and data sharing agreements.

\heading{Author Contributions}

M.S., F.M., and S.J. conceived the study and designed the experiments. 
M.S, H.Q., Y.Y.Y., Y.B., P.M., and J.Y.J.L. curated the pretraining dataset. 
H.Q., Y.Y.Y., P.M., S.L., W.W., P.C., C.N.C., P.S., J.S., J.L.L., H.A.M., A.M., and S.J.R. did the quality control on the pretraining dataset. 
N.E.A., S.M., M.H., D.P., Y.T., G.P.N., W.R.B., J.D.E., T.K.C., N.A., M.B., C.M.S., F.J.T., Y.H.K., J.Y., S.S., B.E.H., L.H.L., and Q.M. were involved in the generation of in-house pretraining and evaluation datasets.
M.S., Y.C., Y.W., L.L.W., T.D., A.V., A.L., D.S., M.Z., S.S., A.Z., M.E.M.G., and H.C. ran experiments. 
M.S., Y.C., Y.W., A.H.S., G.J., M.Y.L., R.J.C., D.S., J.T.C.L., L.P.L., G.G., F.M., and S.J. interpreted the experimental results and provided feedback on the study.
M.S., Y.C., Y.W., C.T., A.H.S., G.J., and C.T. prepared the figures. 
M.S., G.J., C.A.P., and S.J.W. prepared and reviewed the tutorials and code for KRONOS.
M.S., Y.C., A.H.S., G.J., D.S., J.T.C.L., F.M., and S.J. prepared the manuscript with input from all co-authors.
S.J. and F.M. supervised the research.

\heading{Acknowledgements}

We thank members of the Jiang and Mahmood laboratories for valuable discussions and feedback throughout the project. We thank Timothy Janicki at BWH, and Richard Kenny and the system administration staff at the MGB Enterprise Research Infrastructure \& Services (ERIS) Research Computing Core, for their support in maintaining access to computing resources. We thank Drs.\ David Einstein, Steve Balk, Yue Sun, and Matthias Matter for their clinical and pathological expertise. Some illustrations were created with BioRender.com. 

This work was supported by the BWH \& MGH Pathology, BWH President's Fund, Massachusetts Life Sciences Center, National Institute of General Medical Sciences (NIGMS) R35GM138216 (F.M.), and the BWH President's Scholar Fund (G.G.). S.J. is supported in part by the National Institutes of Health (NIH) under awards DP2AI171139, P01AI177687, R01AI149672, and U24CA224331; the Gilead Research Scholars Program in Hematologic Malignancies; a Sanofi iAward; the Bill \& Melinda Gates Foundation (INV-002704); the Broad Institute Next Generation Award; the Dye Family Foundation; and the Bridge Project, a partnership between the Koch Institute for Integrative Cancer Research at MIT and the Dana-Farber/Harvard Cancer Center. Q.M. is supported in part by the NIH under awards R01GM152585, R01DK138504, P01CA278732, and U54AG075931, and the Pelotonia Institute of Immuno-Oncology. T.K.C. is supported in part by the Dana-Farber/Harvard Cancer Center Kidney SPORE (2P50CA101942-16) and Program 5P30CA006516-56, the Kohlberg Chair at Harvard Medical School and the Trust Family, Michael Brigham, Pan Mass Challenge, Hinda and Arthur Marcus Fund and Loker Pinard Funds for Kidney Cancer Research at DFCI. P.S. is supported by a mobility grant from Institut Servier, with research funding provided to the institution by Abbvie and BeiGene. J.L.L. is supported by a National Science Scholarship (PhD) from the Agency for Science, Technology and Research, Singapore (BM/NDR/18/003) and is a Schmidt Science Fellow. J.S. is supported by a Roche Postdoctoral Fellowship.

F.M. and L.P.L. are co-founders and advisors of Modella AI. F.M., R.J.C., M.Y.L., and L.P.L. hold equity interests in Modella AI. F.M. serves on the advisory board for Danaher. S.J. is a co-founder of Elucidate Bio Inc, serves on its Board of Directors and Scientific Advisory Board, and has received research support from Roche and Novartis unrelated to this work. S.J.R. has received research support from Affimed, Merck, and Bristol-Myers Squibb (BMS), serves on the Scientific Advisory Board for Immunitas Therapeutics, and is part of the BMS International Immuno-Oncology Network (II-ON), all unrelated to this work. L.-H.L. and J.-Y.J.L. are co-founders and part-time executives of ImmunoQs Pte. Ltd. All work in this study was performed under BII, A*STAR without any financial or other support from ImmunoQs Pte. Ltd. T.K.C. reports institutional and/or personal support for research, advisory boards, consultancy, and/or honoraria from: Alkermes, Arcus Bio, AstraZeneca, Aravive, Aveo, Bayer, Bristol Myers-Squibb, Bicycle Therapeutics, Calithera, Circle Pharma, Deciphera Pharmaceuticals, Eisai, EMD Serono, Exelixis, GlaxoSmithKline, Gilead, HiberCell, IQVA, Infinity, Institut Servier, Ipsen, Jansen, Kanaph, Lilly, Merck, Nikang, Neomorph, Nuscan/PrecedeBio, Novartis, Oncohost, Pfizer, Roche, Sanofi/Aventis, Scholar Rock, Surface Oncology, Takeda, Tempest, Up-To-Date, and CME providers, all outside the submitted work. T.K.C. holds institutional patents on molecular alterations and immunotherapy response/toxicity, rare genitourinary cancers, and ctDNA/liquid biopsies, and equity in Tempest, Pionyr, Osel, Precede Bio, CureResponse, InnDura Therapeutics, Primium, Abalytics, and Faron Pharma. T.K.C. serves on committees for NCCN, GU Steering Committee, ASCO, ESMO, ACCRU, and KidneyCan. C.M.S. is a co-founder and shareholder of Vicinity Bio GmbH and serves on the scientific advisory board of Enable Medicine, Inc., which has provided research funding, all outside the current work. J.T.C.L. is a co-founder, equity holder, and board member of Alpenglow Biosciences Inc., which has licensed 3D pathology technologies developed in his laboratory. S.S. reports receiving commercial research grants from Bristol-Myers Squibb, AstraZeneca, Exelixis, and Novartis. M.H. is a scientific advisory board member of CellFormatica Inc, unrelated to the submitted work. G.P.N. is a co-founder of Akoya Biosciences and a scientific advisory board member for Akoya Biosciences.

A provisional patent application related to the KRONOS architecture and its applications has been filed with MGB.


\end{spacing}

\newpage
\begin{nolinenumbers}
\section*{References}
\bibliographystyle{nature}
\bibliography{ref}
\end{nolinenumbers}

\newpage

\setcounter{figure}{0}
\renewcommand{\figurename}{Figure}
\renewcommand{\thefigure}{S\arabic{figure}}


\begin{figure}[!ht]
    \centering
    \includegraphics[width=0.8\textwidth]{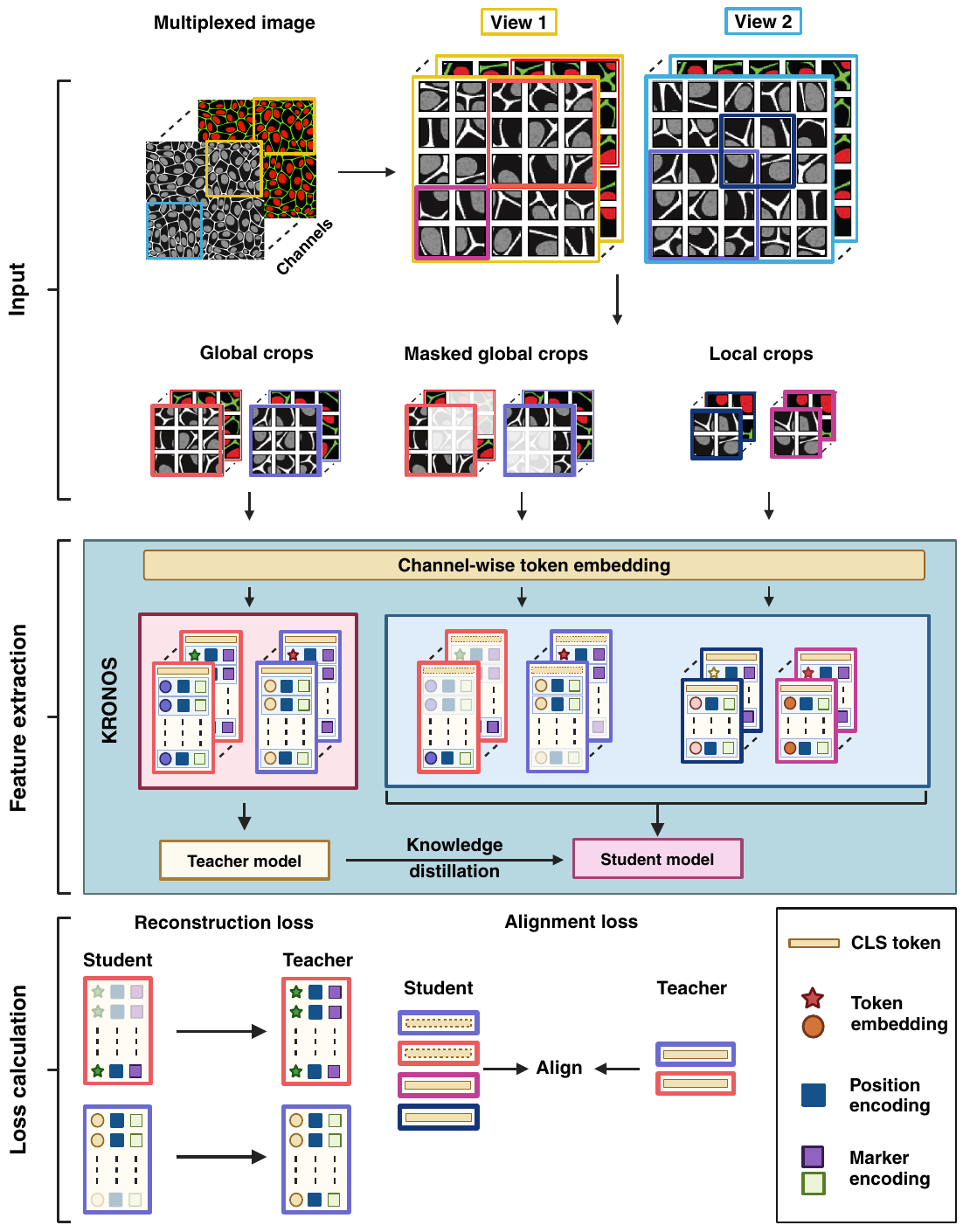}
    \caption{\textbf{Overview of KRONOS pretraining strategy using DINO-v2 adapted for multiplexed imaging encoding.}
    Input multiplexed images are split into channel-wise views, from which global, masked, and local crops are sampled. Each view is processed through a channel-wise token embedding module that incorporates positional, marker, and token-level information. The teacher model receives unmasked crops, while the student model processes masked or local variants. Pretraining is driven by two loss components: (i) a reconstruction loss aligning the student’s masked patch tokens to the teacher’s unmasked representations, and (ii) an alignment loss enforcing consistency between student and teacher CLS tokens. This design extends the DINO-v2 framework to capture spatial and channel-specific structure in multiplexed tissue images.}
    \label{fig:kronos}
\end{figure}

\begin{figure}[!ht]
    \centering
    \includegraphics[width=0.9\textwidth]{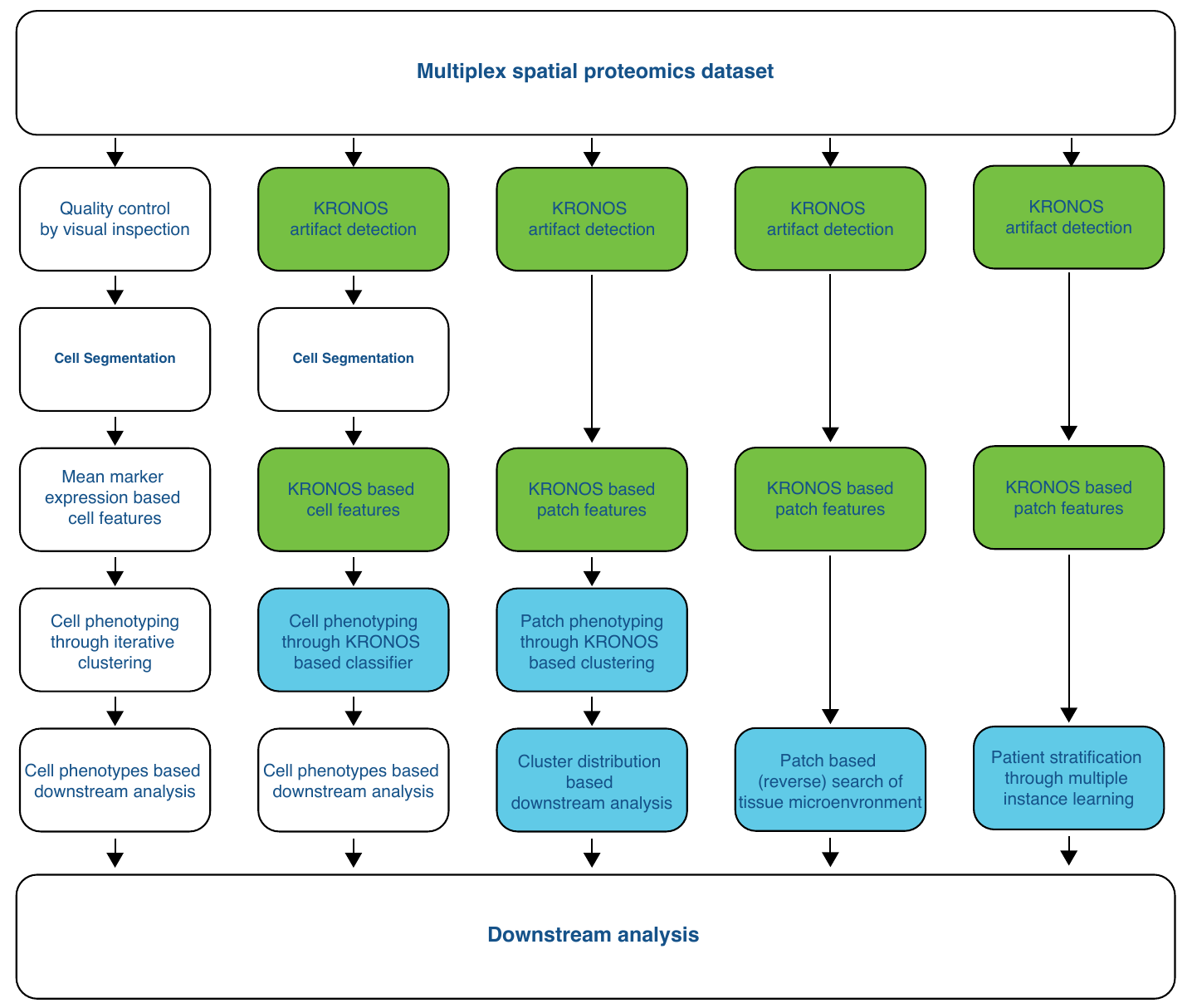}
    \caption{
    \textbf{Comparison of the traditional and KRONOS workflows for analyzing spatial proteomics images.}
    The traditional approach for multiplex imaging analysis relies on visual quality control, extracting mean marker expression per segmented cell, and unsupervised clustering for cell phenotyping. In contrast, KRONOS workflows integrate artifact detection for cell phenotyping, image classification, and weakly-supervised multiple instance learning. Moreover, KRONOS can provide both the cell-based and patch-based features, allowing for downstream analyses at multiple scales.
    }
    \label{fig:kronos_workflow}
\end{figure}

\begin{figure}[!ht]
    \centering
    \includegraphics[width=0.99\textwidth]{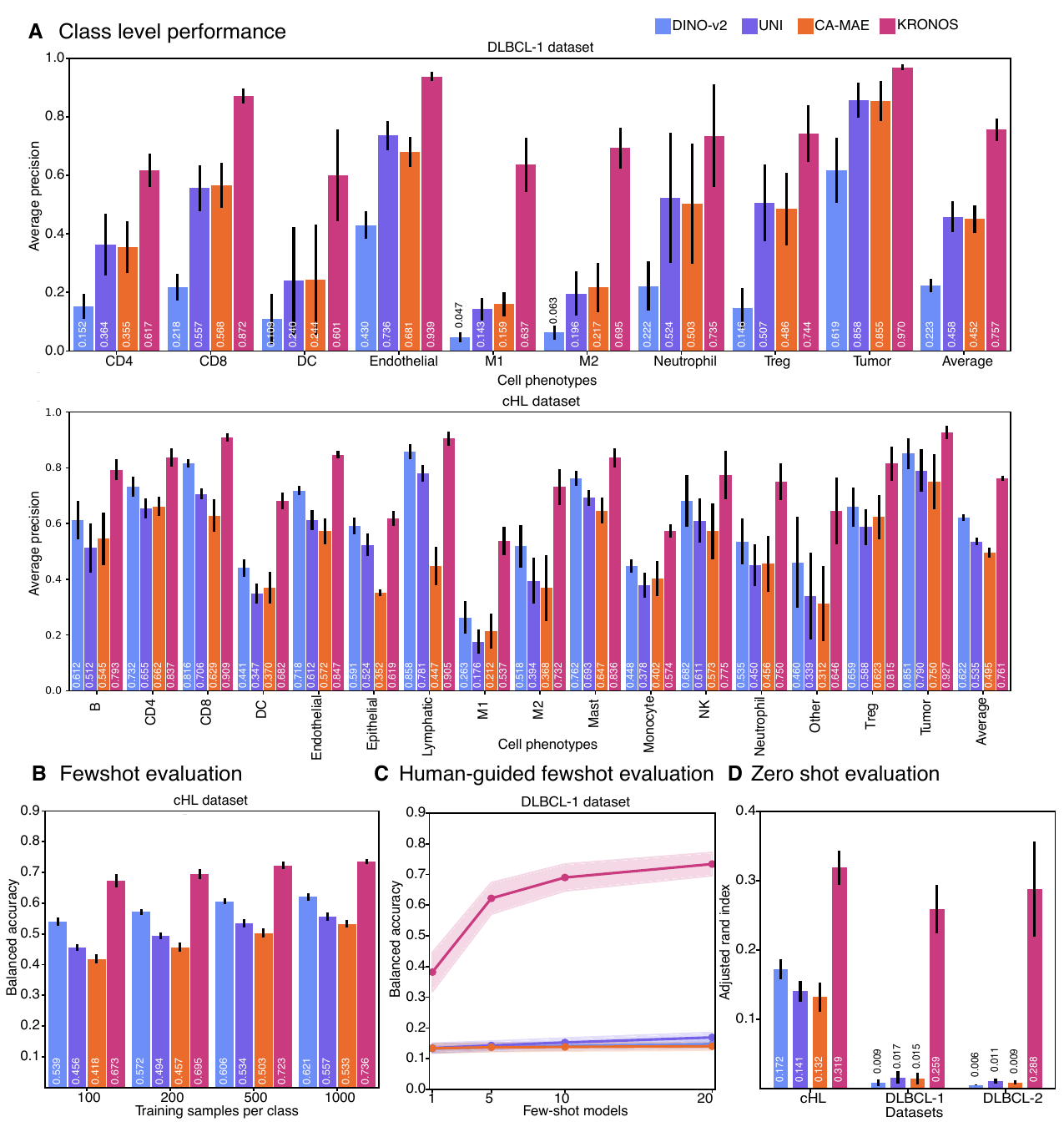}
    \caption{\textbf{Cell phenotyping, few-shot and zero-shot results.} This is an extension of the results in \textbf{Figure~\ref{fig:cell_phenotyping}}. \textbf{A,} Class-wise average precision comparison of KRONOS and baselines on DLBCL-1 and cHL cohorts. \textbf{B,} Few-shot evaluation of KRONOS and baselines using from 100 to 1,000 training examples per class on cHL cohort. \textbf{C,} Human-guided few-shot evaluation of KRONOS on DLBCL-1 cohort, where for a given image, 1 to 20 manual annotations per class are provided as training examples. \textbf{D,} Evaluation of the unsupervised clustering quality of patch embedding space for three cohorts, as measured by adjusted rand index. The number of ground truth clusters (cell types) was 16 for cHL and 9 for DLBCL cohorts. In \textbf{A, B, D}, each bar represents the mean performance, with error bars indicating the standard deviation across four folds. In \textbf{C}, the shade represents the standard deviation across 100 trials and DLBCL-1 images.}
    \label{fig:supp_image_classification}
\end{figure}

\begin{figure}[!ht]
    \centering
    \includegraphics[width=0.99\textwidth]{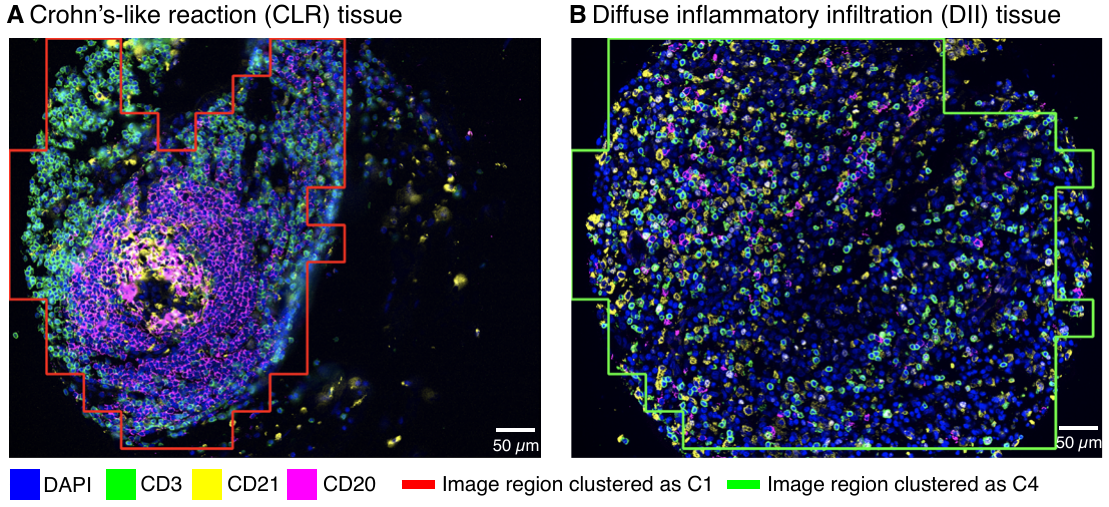}
    
    \caption{\textbf{Clustering of colorectal cancer\cite{schurch2020coordinated} (CRC) immune tumor microenvironments (iTMEs) using KRONOS.} CRC tissue images exhibiting Crohn's-like reaction (CLR) and diffuse inflammatory infiltration (DII) iTMEs were clustered using K-means ($K=8$) on KRONOS embeddings generated from image patches of size $96 \times 96 \mu m$ with 50\% overlap. Clusters C1 and C4 were found to be significantly enriched in CLR and DII regions, respectively.
    \textbf{A}, Representative CLR tissue core with red boundary indicating regions assigned to cluster C1, showing strong spatial overlap with the CLR iTME.
    \textbf{B}, Representative DII tissue core with green boundary indicating regions assigned to cluster C4, showing strong spatial overlap with the DII iTME.
    The observed alignment between these clusters and known iTME types demonstrates that KRONOS embeddings can capture biologically meaningful tissue phenotypes through unsupervised clustering, without the use of ground truth annotations. Colored square boxes indicate the marker-specific color coding used in the images.}
    \label{fig:crc_clustering}
\end{figure}

\begin{figure}[!ht]
    \centering
    \includegraphics[width=0.99\textwidth]{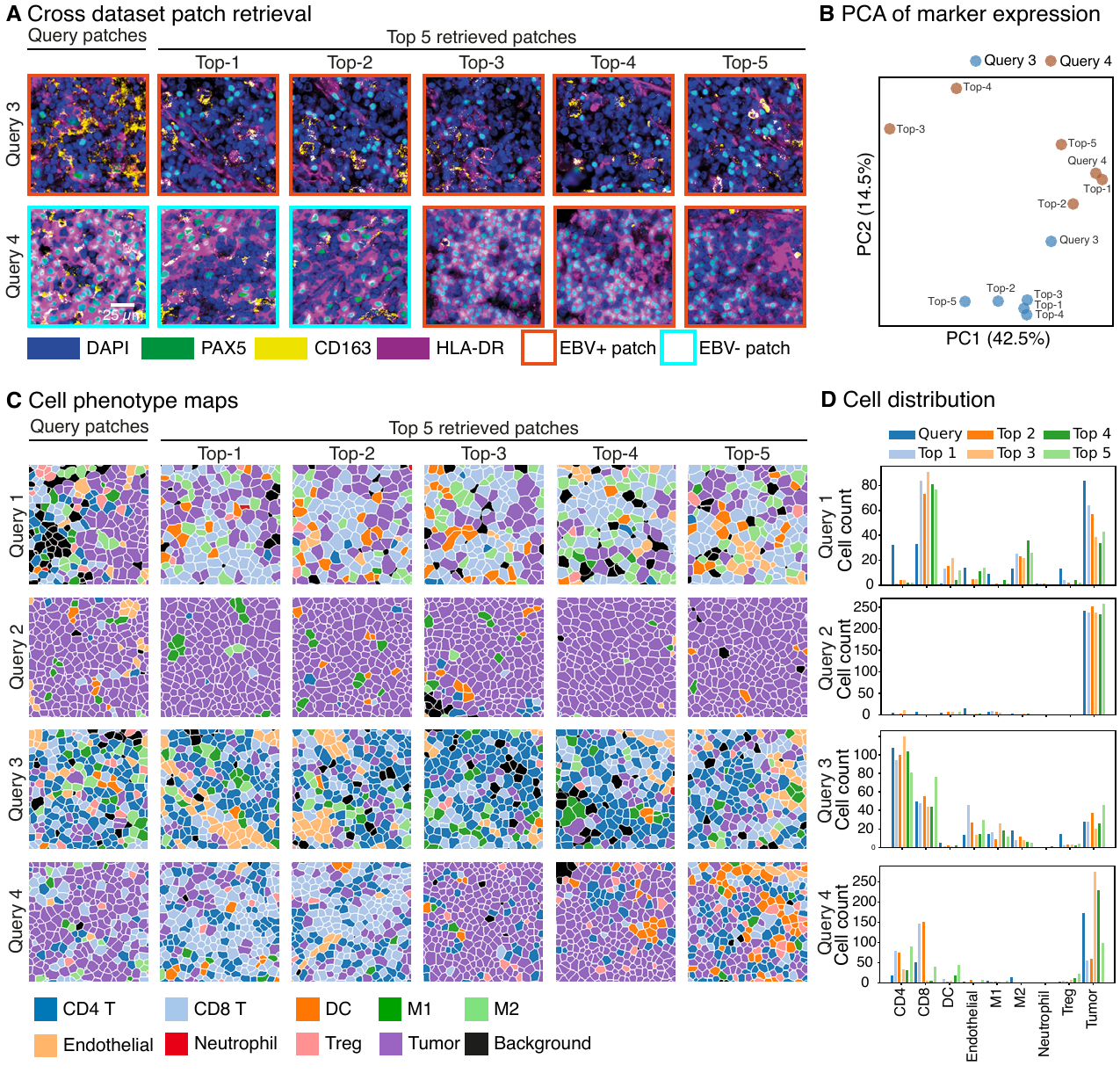}
    \caption{
    \textbf{Reverse search through patch retrieval with KRONOS.} Query 1\&3 patches are from DLBCL2 and Query 2\&4 patches are from DLBCL1.
    \textbf{A}, Two multiplexed patches (Query 3: EBV-positive and Query 4: EBV-negative) and top-5 closest retrieved patches. 
    \textbf{B}, Principal component analysis (PCA) of marker expression profiles for Query 3 and Query 4 and their top 5 retrieved patches. Retrieved patches cluster near their corresponding query, indicating similarity in marker expression.
    \textbf{C}, Cell phenotype maps for four representative query patches and their top 5 retrieved patches, overlaid with ground truth cell types.
    \textbf{D}, Bar plots showing cell type distributions in the query patches and in the top-5 closest retrieved patches.}
    \label{fig:retrieval_supp}
\end{figure}

\begin{figure}[!ht]
    \centering
    \includegraphics[width=0.99\textwidth]{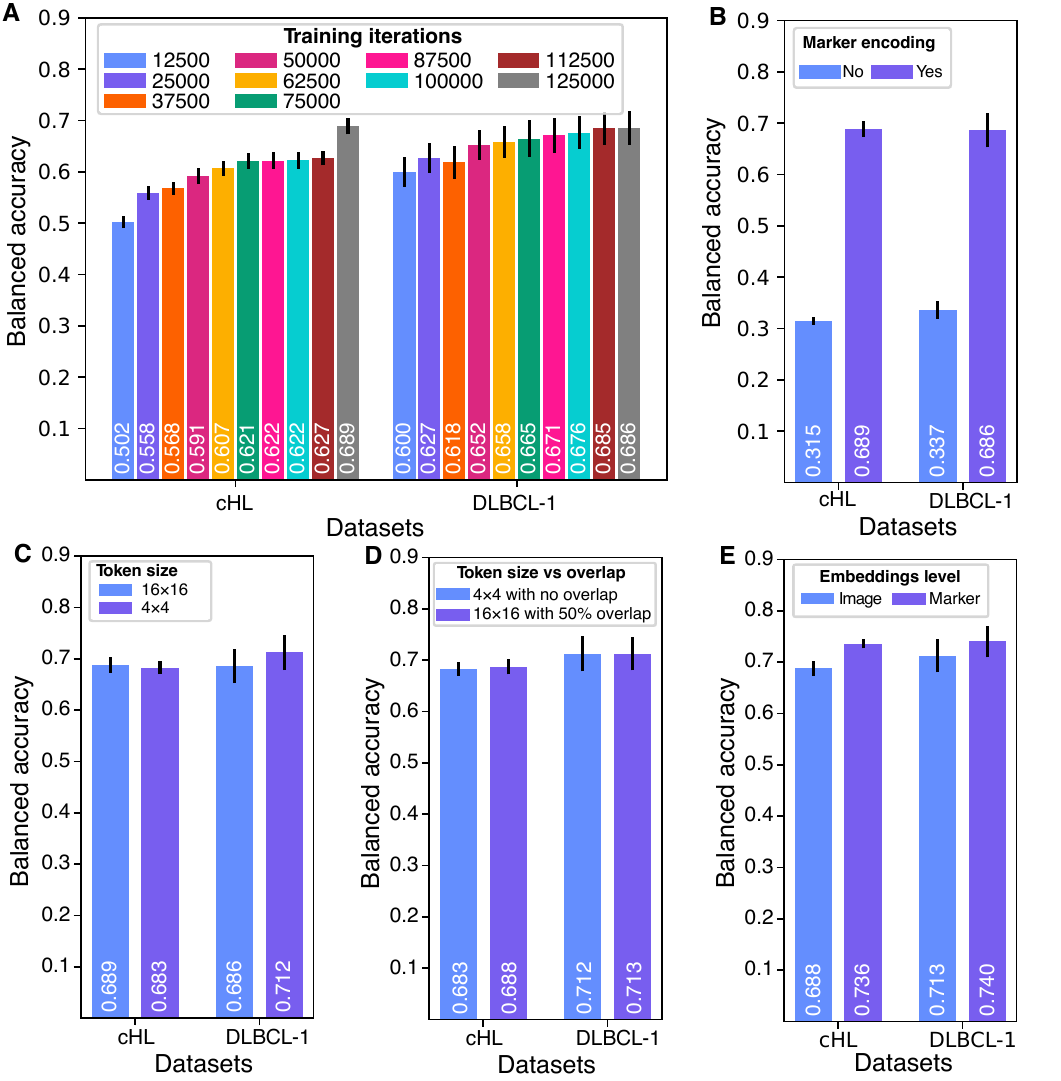}
    \caption{\textbf{Ablation studies of KRONOS.}
    Cell phenotyping measured with balanced accuracy across two datasets (cHL, 16-class classification and DLBCL-1, 9-class classification) for different architectural and design choices in KRONOS. Error bars represent test standard deviation over four folds.
    \textbf{A}, Impact of the number of training iterations during model pretraining.
    \textbf{B}, Effect of including marker encoding.
    \textbf{C}, Impact of token size when pretraining using 4$\times$4-pixel tokens vs. 16$\times$16.  
    \textbf{D}, Comparison between 4$\times$4-pixel tokens vs. 16$\times$16-pixel tokens with 50\% overlap between tokens. 
    \textbf{E}, Effect of using the CLS token embedding (Image) vs. the spatial average of marker embeddings (Marker).
    }
    \label{fig:supp_ablation}
\end{figure}

\begin{figure}[!ht]
    \centering
    \includegraphics{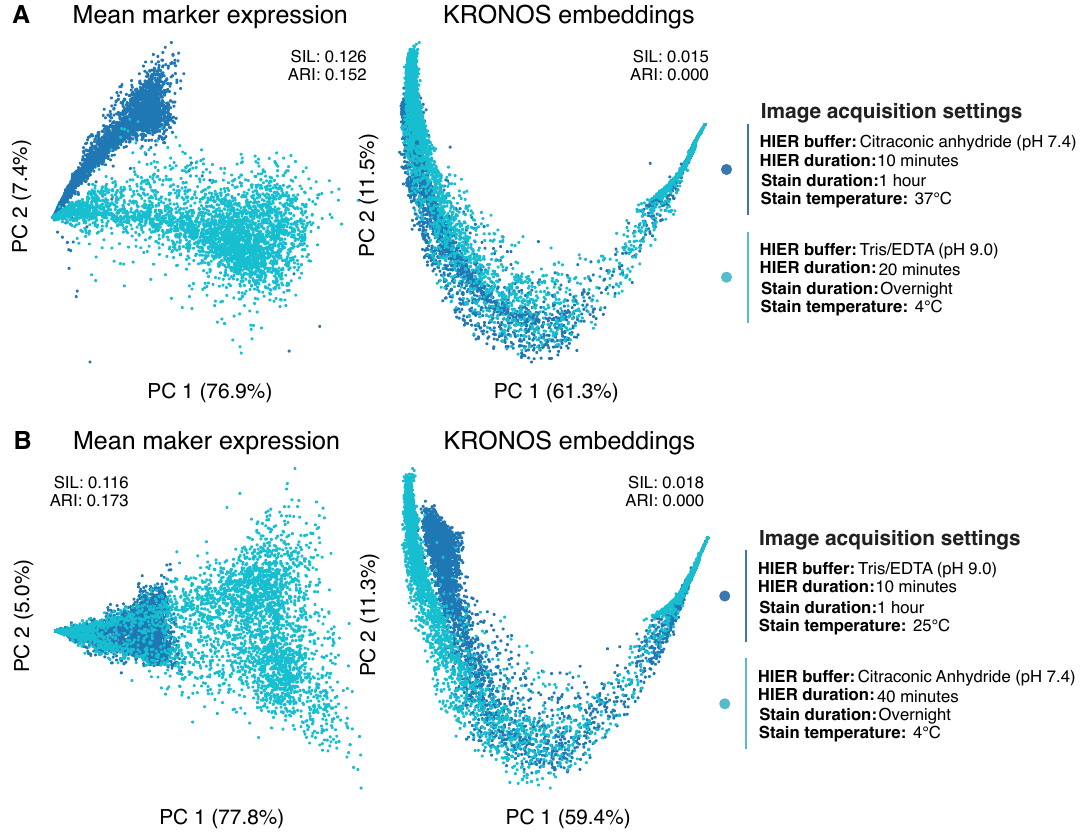}
    \caption{\textbf{Batch effect analysis.}
    \textbf{A\&B}, PCA visualizations of embeddings computed using the \textbf{Mean marker expression} baseline (left) and \textbf{KRONOS} (right) under varying experimental conditions. Each point represents a patch, colored by batch condition. Mean marker expression represents mean intensity of each markers in a patch. The silhouette index (SIL) and adjusted rand index (ARI) are reported for each embedding as quantitative measures of batch-based clustering, where higher values indicate stronger batch effects.}
    \label{fig:batch_effect}
\end{figure}

\begin{figure}[!ht]
    \centering
    \includegraphics[width=0.99\textwidth]{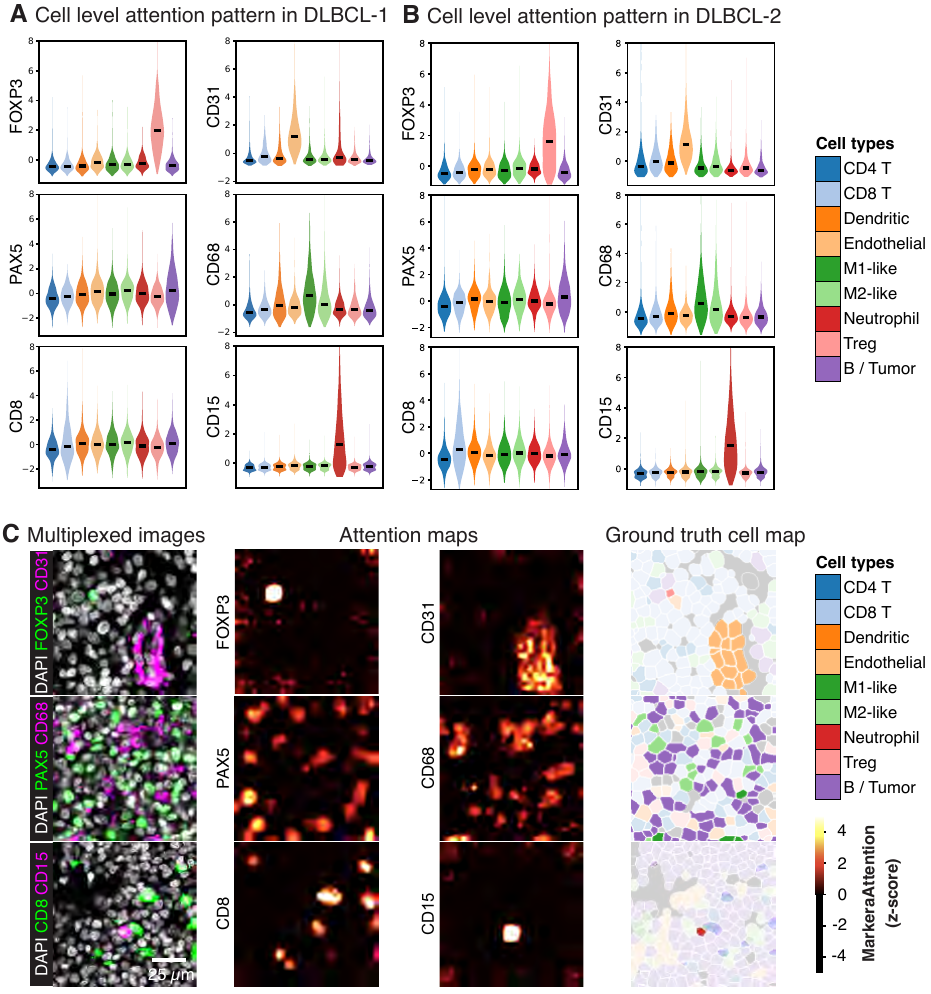}
    \caption{
    \textbf{Analysis of attention scores in KRONOS.}
    \textbf{A,B}, Violin plots showing the distribution of marker attention scores (z-scored) for six representative markers (CD15, CD31, CD68, CD8, FOXP3, and PAX5) across nine cell phenotypes in DLBCL-1 \textbf{A} and DLBCL-2
    \textbf{B}, Each distribution was computed from a randomly selected subset of 1,000 cells per phenotype. Horizontal bars indicate median values. The observed differences highlight marker–phenotype associations learned by KRONOS.
    \textbf{C}, \textit{Multiplex images}: Representative multiplexed images colored with DAPI, FOXP3, CD31, PAX5, CD68, CD8 and CD15 in DLBCL-1. \textit{Attention maps}: KRONOS-derived attention maps for each protein marker. \textit{Ground truth cell map}: Annotated ell phenotype maps focusing on the phenotypes associated with selected markers.}
    \label{fig:marker_attention}
\end{figure}

\clearpage
\setcounter{table}{0}
\renewcommand{\tablename}{Table}
\renewcommand{\thetable}{S\arabic{table}}

\begin{table}[]
\centering
\caption{\textbf{Overview of the SPM-47M dataset used for KRONOS pretraining.} Each patch is of size 256$\times$256-pixel with pixel resolution between 0.37$\mu$m/px and 0.5$\mu$m/px. Marker Patches refer to the number of single-marker patches.}
\label{tab:pretrain_dataset}
\begin{tabular}{lllrrl}
\toprule
\textbf{Data Source} & \textbf{Technology} & \textbf{Tissue Type} & \textbf{Multiplex Patches} & \textbf{Marker Patches} & \textbf{Access} \\
\midrule
ImmunoAtlas\cite{lee2021627}               & COMET               & Lung                 & 36,877                      & 1,370,863                                & Public         \\
ImmunoAtlas\cite{lee2021627}               & COMET               & Nasopharynx          & 21,016                      & 754,986                                  & Partial         \\
ImmunoAtlas\cite{lee2021627}               & CODEX               & Skin                 & 1,793                       & 93,616                                    & Public          \\
ImmunoAtlas\cite{lee2021627}               & COMET               & Colorectal           & 39,812                      & 1,382,640                                  & Public          \\
ImmunoAtlas\cite{lee2021627}               & COMET               & Gastric              & 11,733                      & 351,990                                  & Public          \\
ImmunoAtlas\cite{lee2021627}               & COMET               & Tonsil               & 27,073                      & 1,024,770                                & Public          \\
ImmunoAtlas\cite{lee2021627}               & Vectra              & Breast               & 5,397                       & 32,382                                    & Public          \\
ImmunoAtlas\cite{lee2021627}               & Vectra              & Colorectal           & 1,314                       & 9,198                                  & Public          \\
ImmunoAtlas\cite{lee2021627}               & Vectra              & Lung                 & 17,264                      & 120,848                                  & Partial          \\
ImmunoAtlas\cite{lee2021627}               & Vectra              & Ovary                & 1,700                       & 11,900                                   & Public          \\
BIDMC                & CODEX               & Kidney               & 97,957                      & 2,177,801                                 & Private         \\
BIDMC                & CODEX               & Liver                & 17,829                      & 614,497                                   & Private         \\
BIDMC                & CODEX               & Lymph Node           & 123,509                     & 4,233,663                                & Private         \\
BIDMC                & CODEX               & Mixed                & 24,935                      & 447,254                                  & Private         \\
BIDMC                & CODEX               & Nasal                & 22,019                      & 474,880                                  & Private         \\
BIDMC                & CODEX               & Prostate             & 104,092                     & 1,876,621                                 & Private         \\
BIDMC                & CODEX               & Tonsil               & 260,690                     & 7,593,915                                 & Private         \\
NIH-NIAID\cite{lymphoma}            & CellDive            & Lymph Node          & 1,580,114                    & 13,051,759                                & Public          \\
NIH-NIAID\cite{ibex}              & IBEX                & Jejunum              & 448                        & 10,752                                   & Public          \\
NIH-NIAID\cite{ibex}              & IBEX                & Kidney               & 2,024                       & 30,360                                 & Public          \\
NIH-NIAID\cite{ibex}              & IBEX                & Liver                & 551                        & 11,020                                   & Public          \\
NIH-NIAID\cite{ibex}              & IBEX                & Lymph Node          & 1,765                       & 65,305                                  & Public          \\
NIH-NIAID\cite{ibex}              & IBEX                & Lymph Node          & 20,872                      & 709,648                                  & Public          \\
NIH-NIAID\cite{ibex}              & IBEX                & Skin                 & 314                        & 4,710                                   & Public          \\
NIH-NIAID\cite{ibex}              & IBEX                & Spleen               & 2,714                       & 62,422                                  & Public          \\
NIH-NIAID\cite{ibex}            & MxIF                & Lymph Node          & 906,367                     & 4,080,604                                 & Public          \\
Stanford             & CosMx               & Uterine              & 2,947                       & 153,244                                & Private         \\
Stanford             & Orion               & Tonsil               & 77,865                      & 1,245,840                              & Private         \\
Stanford             & Orion               & Uterine              & 71,499                      & 1,142,984                               & Private         \\
UF\cite{hubmap}                   & CODEX               & Spleen               & 112,626                     & 2,498,370                                 & Public          \\
UF\cite{hubmap}                   & CODEX               & Thymus               & 77,158                      & 1,711,748                                 & Public         \\
\bottomrule
\end{tabular}
\end{table}

\begin{table}[ht]
\centering
\caption{\textbf{List of the 175 markers used for KRONOS pretraining.}}
\label{tab:marker_list}
\begin{minipage}{0.95\textwidth}
\begin{multicols}{4}
\raggedcolumns
\footnotesize
\noindent
a-SMA\\
ACE2\\
ARID1A\\
ASS1\\
ATP5A\\
ATRX\\
B7H3\\
BCL2\\
BCL6\\
BDCA-2\\
b2M\\
b-CATENIN\\
b-TUBULIN\\
C1Q\\
C4A\\
CA9\\
CATHEPSIN L\\
CCR4\\
CCR6\\
CCR7\\
CD1A\\
CD1B\\
CD1C\\
CD10\\
CD103\\
CD106\\
CD107A\\
CD11B\\
CD11C\\
CD134\\
CD138\\
CD14\\
CD15\\
CD16\\
CD162\\
CD163\\
CD164\\
CD19\\
CD2\\
CD20\\
CD206\\
CD21\\
CD23\\
CD25\\
CD27\\
CD271\\
CD28\\
CD3\\
CD30\\
CD31\\
CD34\\
CD35\\
CD36\\
CD38\\
CD39\\
CD4\\
CD40\\
CD44\\
CD45\\
CD45RA\\
CD45RO\\
CD49A\\
CD5\\
CD54\\
CD56\\
CD57\\
CD61\\
CD66B\\
CD68\\
CD69\\
CD7\\
CD8\\
CDT1\\
CGAS\\
CLEAVED CASP3\\
CLEC9A\\
CLUSTERIN\\
COLLAGEN\\
CS\\
CTLA4\\
CXCL13\\
CXCR5\\
CYTOKERATIN\\
DAPI\\
DC-SIGN\\
DESMIN\\
E-CADHERIN\\
EBNA1\\
EGFR\\
EOMES\\
EPCAM\\
ERB\\
FIBRINOGEN\\
FN1\\
FOXP3\\
G6PD\\
GALECTIN-3\\
GATA3\\
GEMININ\\
GITR\\
GLUT1\\
GLYCOPHORIN-A\\
GZMA\\
GZMB\\
H2AX\\
H3K27AC\\
H3K27ME3\\
HEPATOCYTE-AG\\
HER2\\
HEY1\\
HISTONE-H3\\
HLA-1\\
HLA-DR\\
HLA-DRA\\
HLA-DRBPB\\
HSP27\\
IBA1\\
ICAM1\\
ICOS\\
IDH\\
IDO1\\
IGA1\\
IGA2\\
IGD\\
IGM\\
IL-1b\\
INOS\\
IRF4\\
KI-67\\
LAG3\\
LAMINA\\
LANGERIN\\
LMP1\\
LUMICAN\\
LYSOZYME\\
LYVE-1\\
MCT\\
MMP9\\
MPO\\
MUC1\\
MUC5AC\\
NAKATP\\
NOTCH2\\
P53\\
PAX5\\
PD1\\
PD-L1\\
pHH3\\
PODOCIN\\
PODOPLANIN\\
PROX1\\
PTEN\\
RAS\\
RB\\
RELA\\
RELB\\
S100\\
SIGELC-3\\
SIGLEC-7\\
SIGLEC-9\\
SPARC\\
STING\\
TBET\\
TCF1\\
TCR-b\\
TCR-gd\\
TCR-Va7.2\\
TIM3\\
TMPRSS2\\
TOX\\
TREM2\\
UROMODULIN\\
VDAC1\\
VIMENTIN\\
VISTA
\end{multicols}
\end{minipage}
\end{table}

\begin{table}[ht]
\centering
\caption{\textbf{Cell type distribution in cHL, DLBCL-1 and DLBCL-2 used in KRONOS benchmarking.} cHL: Classical Hodgkin Lymphoma, DLBCL-1: Diffuse Large B-Cell Lymphoma-1, DLBCL-2: Diffuse Large B-Cell Lymphoma-2.}
\label{tab:cell_dataset}
\begin{tabular}{lccc}
\toprule
\textbf{Cell Type} & \textbf{cHL} & \textbf{DLBCL-1} & \textbf{DLBCL-2} \\
\midrule
CD4         & 34,566  & 35,313   & 286,329 \\
CD8         & 16,453  & 59,714   & 210,417 \\
DC          & 9,011   & 21,114   & 56,968  \\
Endothelial & 8,287   & 18,981   & 48,422  \\
M1          & 2,944   & 13,371   & 29,286  \\
M2          & 6,997   & 18,181   & 65,038  \\
Neutrophil  & 3,260   & 9,146    & 4,080   \\
Treg        & 3,075   & 6,965    & 32,145  \\
Tumor       & 7,557   & 178,556  & 365,758 \\
B           & 15,064  & —        & —       \\
Epithelial  & 2,176   & —        & —       \\
Lymphatic   & 3,606   & —        & —       \\
Mast        & 3,118   & —        & —       \\
Monocyte    & 6,598   & —        & —       \\
NK          & 6,978   & —        & —       \\
Other       & 4,862   & —        & —       \\
\midrule
\textbf{Total Cells}   & 134,552 & 361,341  & 1,098,443 \\
\textbf{Total Classes} & 16      & 9         & 9         \\
\bottomrule
\end{tabular}
\end{table}

\begin{table}[ht]
\centering
\caption{\textbf{List of markers measured in the cHL, DLBCL-1, and DLBCL-2 datasets.} cHL: Classical Hodgkin Lymphoma, DLBCL-1: Diffuse Large B-Cell Lymphoma-1, DLBCL-2: Diffuse Large B-Cell Lymphoma-2.}
\label{tab:phenotyping_markers}
\begin{tabular}{lll}
\toprule
\textbf{cHL}       & \textbf{DLBCL-1} & \textbf{DLBCL-2} \\
\midrule
DAPI         & DAPI       & DAPI       \\
CD11b        & CD11c      & CD11c      \\
CD11c        & CD15       & CD15       \\
CD15         & CD163      & CD163      \\
CD163        & CD20       & CD20       \\
CD20         & CD3        & CD3        \\
CD206        & CD31       & CD31       \\
CD30         & CD4        & CD4        \\
CD31         & CD68       & CD68       \\
CD4          & CD8        & CD8        \\
CD56         & FoxP3      & FoxP3      \\
CD68         & Pax5       & Pax5       \\
CD7          & —          & —          \\
CD8          & —          & —          \\
Cytokeratin  & —          & —          \\
FoxP3        & —          & —          \\
MCT          & —          & —          \\
Podoplanin   & —          & —          \\
\bottomrule
\end{tabular}
\end{table}

\begin{table}[ht]
\centering
\caption{\textbf{Distribution of cells across four folds in the three cell phenotyping datasets.} cHL: Classical Hodgkin Lymphoma, DLBCL-1: Diffuse Large B-Cell Lymphoma-1, DLBCL-2: Diffuse Large B-Cell Lymphoma-2.}
\label{tab:cell_folds}
\begin{tabular}{lrrrr}
\toprule
\textbf{Dataset} & \textbf{Fold 1} & \textbf{Fold 2} & \textbf{Fold 3} & \textbf{Fold 4} \\
\midrule
cHL      & 41,955  & 39,927  & 23,588  & 29,082  \\
DLBCL-1  & 77,352  & 99,455  & 108,157 & 76,377  \\
DLBCL-2  & 330,923 & 250,021 & 256,913 & 260,586 \\
\bottomrule
\end{tabular}
\end{table}

\begin{table}[ht]
\centering
\caption{\textbf{Cell phenotyping performance of KRONOS against DINO-v2, UNI, and CA-MAE on cHL, DLBCL-1, and DLBCL-2 datasets.} Results are reported as mean $\pm$ standard deviation over four folds. Best results per dataset and metric are in bold.}
\label{tab:cell_results}
\begin{tabular}{llcccc}
\toprule
\textbf{Dataset} & \textbf{Model} & \textbf{F1-Score} & \textbf{Balanced Accuracy} & \textbf{Average Precision} & \textbf{AUROC} \\
\midrule
\multirow{4}{*}{cHL}
& DINO-v2 & 0.5493 $\pm$ 0.0160 & 0.6210 $\pm$ 0.0121 & 0.6217 $\pm$ 0.0119 & 0.9565 $\pm$ 0.0007 \\
& UNI     & 0.4793 $\pm$ 0.0152 & 0.5570 $\pm$ 0.0136 & 0.5348 $\pm$ 0.0136 & 0.9377 $\pm$ 0.0020 \\
& CA-MAE  & 0.4553 $\pm$ 0.0105 & 0.5331 $\pm$ 0.0123 & 0.4950 $\pm$ 0.0181 & 0.9271 $\pm$ 0.0048 \\
& KRONOS  & \textbf{0.6807 $\pm$ 0.0066} & \textbf{0.7358 $\pm$ 0.0089} & \textbf{0.7614 $\pm$ 0.0084} & \textbf{0.9758 $\pm$ 0.0010} \\
\midrule
\multirow{4}{*}{DLBCL-1}
& DINO-v2 & 0.1932 $\pm$ 0.0316 & 0.2664 $\pm$ 0.0201 & 0.2227 $\pm$ 0.0229 & 0.6623 $\pm$ 0.0161 \\
& UNI     & 0.4073 $\pm$ 0.0529 & 0.5077 $\pm$ 0.0333 & 0.4584 $\pm$ 0.0530 & 0.8474 $\pm$ 0.0191 \\
& CA-MAE  & 0.3992 $\pm$ 0.0498 & 0.5041 $\pm$ 0.0314 & 0.4518 $\pm$ 0.0472 & 0.8455 $\pm$ 0.0179 \\
& KRONOS  & \textbf{0.6669 $\pm$ 0.0492} & \textbf{0.7402 $\pm$ 0.0309} & \textbf{0.7567 $\pm$ 0.0392} & \textbf{0.9638 $\pm$ 0.0045} \\
\midrule
\multirow{4}{*}{DLBCL-2}
& DINO-v2 & 0.2045 $\pm$ 0.0077 & 0.2980 $\pm$ 0.0226 & 0.2432 $\pm$ 0.0103 & 0.6938 $\pm$ 0.0194 \\
& UNI     & 0.4295 $\pm$ 0.0164 & 0.5511 $\pm$ 0.0377 & 0.4985 $\pm$ 0.0244 & 0.8759 $\pm$ 0.0190 \\
& CA-MAE  & 0.4231 $\pm$ 0.0185 & 0.5503 $\pm$ 0.0368 & 0.4946 $\pm$ 0.0300 & 0.8748 $\pm$ 0.0193 \\
& KRONOS  & \textbf{0.6912 $\pm$ 0.0162} & \textbf{0.7969 $\pm$ 0.0125} & \textbf{0.8007 $\pm$ 0.0462} & \textbf{0.9759 $\pm$ 0.0023} \\
\bottomrule
\end{tabular}
\end{table}

\begin{table}[ht]
\centering
\caption{\textbf{Per-class precision performance of KRONOS, DINO-v2, UNI, and CA-MAE on the cHL dataset.} Performance reported as mean $\pm$ standard deviation across four folds. Best model per class is shown in bold.}
\label{tab:cell_class_cHL}
\begin{tabular}{lcccc}
\toprule
\textbf{Cell Type} & \textbf{DINO-v2} & \textbf{UNI} & \textbf{CA-MAE} & \textbf{KRONOS} \\
\midrule
B           & 0.6122 $\pm$ 0.0699 & 0.5122 $\pm$ 0.0871 & 0.5454 $\pm$ 0.0950 & \textbf{0.7927 $\pm$ 0.0375} \\
CD4         & 0.7319 $\pm$ 0.0353 & 0.6553 $\pm$ 0.0359 & 0.6615 $\pm$ 0.0354 & \textbf{0.8369 $\pm$ 0.0322} \\
CD8         & 0.8155 $\pm$ 0.0143 & 0.7059 $\pm$ 0.0197 & 0.6287 $\pm$ 0.0582 & \textbf{0.9089 $\pm$ 0.0141} \\
DC          & 0.4405 $\pm$ 0.0311 & 0.3475 $\pm$ 0.0357 & 0.3698 $\pm$ 0.0574 & \textbf{0.6817 $\pm$ 0.0310} \\
Endothelial & 0.7183 $\pm$ 0.0175 & 0.6123 $\pm$ 0.0346 & 0.5720 $\pm$ 0.0461 & \textbf{0.8473 $\pm$ 0.0129} \\
Epithelial  & 0.5911 $\pm$ 0.0309 & 0.5236 $\pm$ 0.0400 & 0.3516 $\pm$ 0.0109 & \textbf{0.6186 $\pm$ 0.0271} \\
Lymphatic   & 0.8584 $\pm$ 0.0274 & 0.7811 $\pm$ 0.0300 & 0.4469 $\pm$ 0.0691 & \textbf{0.9053 $\pm$ 0.0252} \\
M1          & 0.2629 $\pm$ 0.0575 & 0.1762 $\pm$ 0.0437 & 0.2125 $\pm$ 0.0630 & \textbf{0.5368 $\pm$ 0.0511} \\
M2          & 0.5182 $\pm$ 0.0751 & 0.3942 $\pm$ 0.0822 & 0.3683 $\pm$ 0.1182 & \textbf{0.7317 $\pm$ 0.0648} \\
Mast        & 0.7624 $\pm$ 0.0258 & 0.6929 $\pm$ 0.0286 & 0.6468 $\pm$ 0.0460 & \textbf{0.8357 $\pm$ 0.0333} \\
Monocyte    & 0.4482 $\pm$ 0.0241 & 0.3779 $\pm$ 0.0454 & 0.4025 $\pm$ 0.0616 & \textbf{0.5737 $\pm$ 0.0248} \\
NK          & 0.6818 $\pm$ 0.0920 & 0.6106 $\pm$ 0.0788 & 0.5727 $\pm$ 0.1007 & \textbf{0.7747 $\pm$ 0.0871} \\
Neutrophil  & 0.5354 $\pm$ 0.0825 & 0.4497 $\pm$ 0.0752 & 0.4563 $\pm$ 0.1007 & \textbf{0.7505 $\pm$ 0.0651} \\
Other       & 0.4603 $\pm$ 0.1626 & 0.3393 $\pm$ 0.1562 & 0.3118 $\pm$ 0.1350 & \textbf{0.6458 $\pm$ 0.1205} \\
Treg        & 0.6590 $\pm$ 0.0706 & 0.5877 $\pm$ 0.0645 & 0.6233 $\pm$ 0.0805 & \textbf{0.8150 $\pm$ 0.0606} \\
Tumor       & 0.8512 $\pm$ 0.0558 & 0.7903 $\pm$ 0.0762 & 0.7498 $\pm$ 0.0992 & \textbf{0.9267 $\pm$ 0.0230} \\
\midrule
\textbf{Average} & 0.6217 $\pm$ 0.0119 & 0.5348 $\pm$ 0.0136 & 0.4950 $\pm$ 0.0181 & \textbf{0.7614 $\pm$ 0.0084} \\
\bottomrule
\end{tabular}
\end{table}

\begin{table}[ht]
\centering
\caption{\textbf{Per-class precision performance of KRONOS, DINO-v2, UNI, and CA-MAE on the DLBCL-1 dataset.} Performance reported as mean $\pm$ standard deviation across four folds. Best model per class is shown in bold.}
\label{tab:cell_class_dlbcl_1}
\begin{tabular}{lrrrr}
\toprule
\textbf{Cell Type} & \textbf{DINO-v2} & \textbf{UNI} & \textbf{CA-MAE} & \textbf{KRONOS} \\
\midrule
CD4         & 0.1519 $\pm$ 0.0431 & 0.3639 $\pm$ 0.1054 & 0.3548 $\pm$ 0.0884 & \textbf{0.6174 $\pm$ 0.0560} \\
CD8         & 0.2177 $\pm$ 0.0453 & 0.5568 $\pm$ 0.0777 & 0.5676 $\pm$ 0.0772 & \textbf{0.8721 $\pm$ 0.0256} \\
DC          & 0.1090 $\pm$ 0.0866 & 0.2398 $\pm$ 0.1833 & 0.2438 $\pm$ 0.1878 & \textbf{0.6015 $\pm$ 0.1570} \\
Endothelial & 0.4301 $\pm$ 0.0475 & 0.7362 $\pm$ 0.0489 & 0.6808 $\pm$ 0.0523 & \textbf{0.9389 $\pm$ 0.0164} \\
M1          & 0.0465 $\pm$ 0.0162 & 0.1434 $\pm$ 0.0388 & 0.1595 $\pm$ 0.0402 & \textbf{0.6366 $\pm$ 0.0921} \\
M2          & 0.0629 $\pm$ 0.0240 & 0.1964 $\pm$ 0.0762 & 0.2166 $\pm$ 0.0838 & \textbf{0.6946 $\pm$ 0.0701} \\
Neutrophil  & 0.2222 $\pm$ 0.0835 & 0.5241 $\pm$ 0.2224 & 0.5028 $\pm$ 0.2049 & \textbf{0.7354 $\pm$ 0.1761} \\
Treg        & 0.1456 $\pm$ 0.0695 & 0.5075 $\pm$ 0.1311 & 0.4857 $\pm$ 0.1243 & \textbf{0.7442 $\pm$ 0.0974} \\
Tumor       & 0.6187 $\pm$ 0.1110 & 0.8576 $\pm$ 0.0612 & 0.8547 $\pm$ 0.0690 & \textbf{0.9700 $\pm$ 0.0094} \\
\midrule
\textbf{Average} & 0.2227 $\pm$ 0.0229 & 0.4584 $\pm$ 0.0530 & 0.4518 $\pm$ 0.0472 & \textbf{0.7567 $\pm$ 0.0392} \\
\bottomrule
\end{tabular}
\end{table}

\begin{table}[ht]
\centering
\caption{\textbf{Per-class precision performance of KRONOS, DINO-v2, UNI, and CA-MAE on the DLBCL-2 dataset.} Performance reported as mean $\pm$ standard deviation across four folds. Best model per class is shown in bold.}
\label{tab:cell_class_dlbcl_2}
\begin{tabular}{lrrrr}
\toprule
\textbf{Cell Type} & \textbf{DINO-v2} & \textbf{UNI} & \textbf{CA-MAE} & \textbf{KRONOS} \\
\midrule
CD4         & 0.3726 $\pm$ 0.1284 & 0.6433 $\pm$ 0.1418 & 0.6559 $\pm$ 0.1300 & \textbf{0.8452 $\pm$ 0.0687} \\
CD8         & 0.2343 $\pm$ 0.1136 & 0.4833 $\pm$ 0.1288 & 0.5252 $\pm$ 0.1206 & \textbf{0.8476 $\pm$ 0.0477} \\
DC          & 0.0819 $\pm$ 0.0278 & 0.2651 $\pm$ 0.0906 & 0.2705 $\pm$ 0.0860 & \textbf{0.6756 $\pm$ 0.0263} \\
Endothelial & 0.3625 $\pm$ 0.0218 & 0.6783 $\pm$ 0.0062 & 0.6109 $\pm$ 0.0187 & \textbf{0.9122 $\pm$ 0.0197} \\
M1          & 0.0363 $\pm$ 0.0068 & 0.1687 $\pm$ 0.0712 & 0.1889 $\pm$ 0.0831 & \textbf{0.5514 $\pm$ 0.0595} \\
M2          & 0.0834 $\pm$ 0.0356 & 0.2849 $\pm$ 0.0715 & 0.3121 $\pm$ 0.0741 & \textbf{0.7052 $\pm$ 0.0521} \\
Neutrophil  & 0.1958 $\pm$ 0.0557 & 0.4462 $\pm$ 0.1050 & 0.4215 $\pm$ 0.1307 & \textbf{0.8072 $\pm$ 0.0886} \\
Treg        & 0.2900 $\pm$ 0.1022 & 0.7119 $\pm$ 0.0842 & 0.6706 $\pm$ 0.1039 & \textbf{0.9060 $\pm$ 0.0281} \\
Tumor       & 0.5316 $\pm$ 0.1664 & 0.8046 $\pm$ 0.1327 & 0.7958 $\pm$ 0.1417 & \textbf{0.9561 $\pm$ 0.0255} \\
\midrule
\textbf{Average} & 0.2432 $\pm$ 0.0103 & 0.4985 $\pm$ 0.0244 & 0.4946 $\pm$ 0.0300 & \textbf{0.8007 $\pm$ 0.0462} \\
\bottomrule
\end{tabular}
\end{table}

\begin{table}[ht]
\centering
\footnotesize
\caption{\textbf{Generalization experiments of KRONOS for cell phenotyping against DINO-v2, UNI, and CA-MAE.} Models are trained on one dataset and tested on another. Results are reported as mean $\pm$ standard deviation across four folds. Best scores per row are in bold.}
\label{tab:cross_dataset}
\begin{tabular}{lllcccc}
\toprule
\textbf{Train Dataset} & \textbf{Test Dataset} & \textbf{Model} & \textbf{F1-Score} & \textbf{Balanced Acc.} & \textbf{Avg. Precision} & \textbf{AUROC} \\
\midrule
\multirow{4}{*}{DLBCL-1} & \multirow{4}{*}{DLBCL-2} 
  & DINO-v2 & 0.2081 $\pm$ 0.0024 & 0.2928 $\pm$ 0.0022 & 0.2470 $\pm$ 0.0029 & 0.6869 $\pm$ 0.0026 \\
  &        & UNI     & 0.4553 $\pm$ 0.0088 & 0.5629 $\pm$ 0.0108 & 0.5245 $\pm$ 0.0103 & 0.8791 $\pm$ 0.0061 \\
  &        & CA-MAE  & 0.4522 $\pm$ 0.0057 & 0.5651 $\pm$ 0.0074 & 0.5287 $\pm$ 0.0070 & 0.8821 $\pm$ 0.0043 \\
  &        & KRONOS  & \textbf{0.6699 $\pm$ 0.0091} & \textbf{0.7896 $\pm$ 0.0072} & \textbf{0.7803 $\pm$ 0.0050} & \textbf{0.9715 $\pm$ 0.0012} \\
\midrule
\multirow{4}{*}{DLBCL-2} & \multirow{4}{*}{DLBCL-1} 
  & DINO-v2 & 0.2046 $\pm$ 0.0046 & 0.2632 $\pm$ 0.0021 & 0.2152 $\pm$ 0.0036 & 0.6635 $\pm$ 0.0020 \\
  &        & UNI     & 0.4333 $\pm$ 0.0073 & 0.5164 $\pm$ 0.0074 & 0.4626 $\pm$ 0.0046 & 0.8514 $\pm$ 0.0034 \\
  &        & CA-MAE  & 0.4206 $\pm$ 0.0107 & 0.5116 $\pm$ 0.0086 & 0.4504 $\pm$ 0.0106 & 0.8448 $\pm$ 0.0058 \\
  &        & KRONOS  & \textbf{0.7202 $\pm$ 0.0123} & \textbf{0.7505 $\pm$ 0.0100} & \textbf{0.7983 $\pm$ 0.0081} & \textbf{0.9641 $\pm$ 0.0017} \\
\bottomrule
\end{tabular}
\end{table}

\begin{table}[ht]
\centering
\small
\caption{\textbf{Few-shot cell phenotyping of KRONOS against DINO-v2, UNI, and CA-MAE on the cHL dataset.} Training cells refer to the number of training samples per cell type. Results are reported as mean $\pm$ standard deviation across four folds. Bold indicates the best performance per row.}
\label{tab:limited_dataset_chl}
\begin{tabular}{lllllll}
\toprule
\textbf{Training Cells} & \textbf{Model} & \textbf{F1-Score} & \textbf{Balanced Acc.} & \textbf{Avg. Precision} & \textbf{AUROC} \\
\midrule
\multirow{4}{*}{100}
& DINO-v2 & $0.4662 \pm 0.0130$ & $0.5395 \pm 0.0130$ & $0.5250 \pm 0.0129$ & $0.9344 \pm 0.0010$ \\
& UNI     & $0.3798 \pm 0.0120$ & $0.4563 \pm 0.0111$ & $0.4141 \pm 0.0167$ & $0.9001 \pm 0.0024$ \\
& CA-MAE  & $0.3481 \pm 0.0150$ & $0.4180 \pm 0.0149$ & $0.3679 \pm 0.0154$ & $0.8822 \pm 0.0078$ \\
& \textbf{KRONOS}  & $\mathbf{0.6235 \pm 0.0228}$ & $\mathbf{0.6735 \pm 0.0216}$ & $\mathbf{0.7022 \pm 0.0136}$ & $\mathbf{0.9657 \pm 0.0013}$ \\
\midrule
\multirow{4}{*}{200}
& DINO-v2 & $0.4957 \pm 0.0092$ & $0.5718 \pm 0.0088$ & $0.5603 \pm 0.0115$ & $0.9436 \pm 0.0007$ \\
& UNI     & $0.4126 \pm 0.0099$ & $0.4943 \pm 0.0116$ & $0.4613 \pm 0.0129$ & $0.9163 \pm 0.0014$ \\
& CA-MAE  & $0.3855 \pm 0.0118$ & $0.4570 \pm 0.0150$ & $0.4141 \pm 0.0151$ & $0.9007 \pm 0.0068$ \\
& \textbf{KRONOS}  & $\mathbf{0.6454 \pm 0.0153}$ & $\mathbf{0.6950 \pm 0.0166}$ & $\mathbf{0.7282 \pm 0.0077}$ & $\mathbf{0.9700 \pm 0.0013}$ \\
\midrule
\multirow{4}{*}{500}
& DINO-v2 & $0.5319 \pm 0.0116$ & $0.6057 \pm 0.0095$ & $0.6009 \pm 0.0128$ & $0.9524 \pm 0.0008$ \\
& UNI     & $0.4544 \pm 0.0116$ & $0.5342 \pm 0.0134$ & $0.5072 \pm 0.0159$ & $0.9301 \pm 0.0018$ \\
& CA-MAE  & $0.4290 \pm 0.0119$ & $0.5026 \pm 0.0153$ & $0.4590 \pm 0.0187$ & $0.9164 \pm 0.0061$ \\
& \textbf{KRONOS}  & $\mathbf{0.6692 \pm 0.0082}$ & $\mathbf{0.7232 \pm 0.0119}$ & $\mathbf{0.7489 \pm 0.0086}$ & $\mathbf{0.9737 \pm 0.0017}$ \\
\midrule
\multirow{4}{*}{1,000}
& DINO-v2 & $0.5493 \pm 0.0160$ & $0.6210 \pm 0.0121$ & $0.6217 \pm 0.0119$ & $0.9565 \pm 0.0007$ \\
& UNI     & $0.4793 \pm 0.0152$ & $0.5570 \pm 0.0136$ & $0.5348 \pm 0.0136$ & $0.9377 \pm 0.0020$ \\
& CA-MAE  & $0.4553 \pm 0.0105$ & $0.5331 \pm 0.0123$ & $0.4950 \pm 0.0181$ & $0.9271 \pm 0.0048$ \\
& \textbf{KRONOS}  & $\mathbf{0.6807 \pm 0.0066}$ & $\mathbf{0.7358 \pm 0.0089}$ & $\mathbf{0.7614 \pm 0.0084}$ & $\mathbf{0.9758 \pm 0.0010}$ \\
\bottomrule
\end{tabular}
\end{table}

\begin{table}[ht]
\centering
\small
\caption{\textbf{Few-shot cell phenotyping of KRONOS against DINO-v2, UNI, and CA-MAE on the DLBCL-1 dataset.} Training cells refers to the number of training samples per cell type. Results are reported as mean $\pm$ standard deviation across XX folds. Bold indicates the best performance per row.}
\label{tab:limited_dataset_dlbcl1}
\begin{tabular}{lllllll}
\toprule
\textbf{Training Cells} & \textbf{Model} & \textbf{F1-Score} & \textbf{Balanced Acc.} & \textbf{Avg. Precision} & \textbf{AUROC} \\
\midrule
\multirow{4}{*}{100}
& DINO-v2 & $0.1038 \pm 0.0087$ & $0.1396 \pm 0.0046$ & $0.1239 \pm 0.0008$ & $0.5348 \pm 0.0020$ \\
& UNI     & $0.1761 \pm 0.0251$ & $0.2366 \pm 0.0085$ & $0.1948 \pm 0.0124$ & $0.6319 \pm 0.0073$ \\
& CA-MAE  & $0.1046 \pm 0.0098$ & $0.1428 \pm 0.0019$ & $0.1235 \pm 0.0025$ & $0.5395 \pm 0.0054$ \\
& \textbf{KRONOS} & $\mathbf{0.6395 \pm 0.0540}$ & $\mathbf{0.7143 \pm 0.0410}$ & $\mathbf{0.7318 \pm 0.0532}$ & $\mathbf{0.9576 \pm 0.0086}$ \\
\midrule
\multirow{4}{*}{200}
& DINO-v2 & $0.1190 \pm 0.0138$ & $0.1584 \pm 0.0068$ & $0.1397 \pm 0.0040$ & $0.5564 \pm 0.0070$ \\
& UNI     & $0.2314 \pm 0.0332$ & $0.3080 \pm 0.0205$ & $0.2537 \pm 0.0228$ & $0.6965 \pm 0.0156$ \\
& CA-MAE  & $0.1522 \pm 0.0239$ & $0.2079 \pm 0.0134$ & $0.1598 \pm 0.0128$ & $0.6057 \pm 0.0117$ \\
& \textbf{KRONOS} & $\mathbf{0.6490 \pm 0.0550}$ & $\mathbf{0.7276 \pm 0.0361}$ & $\mathbf{0.7445 \pm 0.0498}$ & $\mathbf{0.9604 \pm 0.0073}$ \\
\midrule
\multirow{4}{*}{500}
& DINO-v2 & $0.1487 \pm 0.0183$ & $0.2053 \pm 0.0073$ & $0.1692 \pm 0.0111$ & $0.6034 \pm 0.0083$ \\
& UNI     & $0.3314 \pm 0.0474$ & $0.4238 \pm 0.0300$ & $0.3694 \pm 0.0470$ & $0.7884 \pm 0.0215$ \\
& CA-MAE  & $0.2897 \pm 0.0442$ & $0.3800 \pm 0.0316$ & $0.3177 \pm 0.0401$ & $0.7567 \pm 0.0226$ \\
& \textbf{KRONOS} & $\mathbf{0.6590 \pm 0.0496}$ & $\mathbf{0.7369 \pm 0.0310}$ & $\mathbf{0.7543 \pm 0.0419}$ & $\mathbf{0.9630 \pm 0.0054}$ \\
\midrule
\multirow{4}{*}{1000}
& DINO-v2 & $0.1932 \pm 0.0316$ & $0.2664 \pm 0.0201$ & $0.2227 \pm 0.0229$ & $0.6623 \pm 0.0161$ \\
& UNI     & $0.4073 \pm 0.0529$ & $0.5077 \pm 0.0333$ & $0.4584 \pm 0.0530$ & $0.8474 \pm 0.0191$ \\
& CA-MAE  & $0.3992 \pm 0.0498$ & $0.5041 \pm 0.0314$ & $0.4518 \pm 0.0472$ & $0.8455 \pm 0.0179$ \\
& \textbf{KRONOS} & $\mathbf{0.6669 \pm 0.0492}$ & $\mathbf{0.7402 \pm 0.0309}$ & $\mathbf{0.7567 \pm 0.0392}$ & $\mathbf{0.9638 \pm 0.0045}$ \\
\bottomrule
\end{tabular}
\end{table}

\begin{table}[ht]
\centering
\small
\caption{\textbf{Few-shot cell phenotyping of KRONOS against DINO-v2, UNI, and CA-MAE on the DLBCL-2 dataset.} Training cells refers to the number of training samples per cell type. Results are reported as mean $\pm$ standard deviation across XX folds. Bold indicates the best performance per row.}
\label{tab:limited_dataset_dlbcl2}
\begin{tabular}{lllllll}
\toprule
\textbf{Training Cells} & \textbf{Model} & \textbf{F1-Score} & \textbf{Balanced Acc.} & \textbf{Avg. Precision} & \textbf{AUROC} \\
\midrule
\multirow{4}{*}{100}
& DINO-v2 & $0.1090 \pm 0.0057$ & $0.1527 \pm 0.0061$ & $0.1257 \pm 0.0029$ & $0.5568 \pm 0.0079$ \\
& UNI     & $0.1836 \pm 0.0093$ & $0.2598 \pm 0.0108$ & $0.2019 \pm 0.0113$ & $0.6573 \pm 0.0110$ \\
& CA-MAE  & $0.1085 \pm 0.0076$ & $0.1444 \pm 0.0160$ & $0.1228 \pm 0.0037$ & $0.5474 \pm 0.0163$ \\
& \textbf{KRONOS} & $\mathbf{0.6652 \pm 0.0203}$ & $\mathbf{0.7651 \pm 0.0194}$ & $\mathbf{0.7799 \pm 0.0551}$ & $\mathbf{0.9705 \pm 0.0032}$ \\
\midrule
\multirow{4}{*}{200}
& DINO-v2 & $0.1179 \pm 0.0064$ & $0.1732 \pm 0.0089$ & $0.1360 \pm 0.0037$ & $0.5769 \pm 0.0080$ \\
& UNI     & $0.2511 \pm 0.0162$ & $0.3461 \pm 0.0294$ & $0.2829 \pm 0.0146$ & $0.7332 \pm 0.0229$ \\
& CA-MAE  & $0.1616 \pm 0.0081$ & $0.2338 \pm 0.0143$ & $0.1676 \pm 0.0095$ & $0.6337 \pm 0.0137$ \\
& \textbf{KRONOS} & $\mathbf{0.6767 \pm 0.0214}$ & $\mathbf{0.7809 \pm 0.0157}$ & $\mathbf{0.7924 \pm 0.0494}$ & $\mathbf{0.9730 \pm 0.0027}$ \\
\midrule
\multirow{4}{*}{500}
& DINO-v2 & $0.1521 \pm 0.0054$ & $0.2269 \pm 0.0137$ & $0.1742 \pm 0.0078$ & $0.6285 \pm 0.0141$ \\
& UNI     & $0.3544 \pm 0.0186$ & $0.4683 \pm 0.0347$ & $0.4092 \pm 0.0205$ & $0.8239 \pm 0.0230$ \\
& CA-MAE  & $0.3115 \pm 0.0172$ & $0.4263 \pm 0.0328$ & $0.3522 \pm 0.0196$ & $0.7939 \pm 0.0235$ \\
& \textbf{KRONOS} & $\mathbf{0.6799 \pm 0.0258}$ & $\mathbf{0.7918 \pm 0.0205}$ & $\mathbf{0.7981 \pm 0.0484}$ & $\mathbf{0.9751 \pm 0.0036}$ \\
\midrule
\multirow{4}{*}{1000}
& DINO-v2 & $0.2045 \pm 0.0077$ & $0.2980 \pm 0.0226$ & $0.2432 \pm 0.0103$ & $0.6938 \pm 0.0194$ \\
& UNI     & $0.4295 \pm 0.0164$ & $0.5511 \pm 0.0377$ & $0.4985 \pm 0.0244$ & $0.8759 \pm 0.0190$ \\
& CA-MAE  & $0.4231 \pm 0.0185$ & $0.5503 \pm 0.0368$ & $0.4946 \pm 0.0300$ & $0.8748 \pm 0.0193$ \\
& \textbf{KRONOS} & $\mathbf{0.6912 \pm 0.0162}$ & $\mathbf{0.7969 \pm 0.0125}$ & $\mathbf{0.8007 \pm 0.0462}$ & $\mathbf{0.9759 \pm 0.0023}$ \\
\bottomrule
\end{tabular}
\end{table}

\begin{table}[ht]
\centering
\caption{\textbf{Image-wise few-shot performance of KRONOS, DINO-v2, UNI, and CA-MAE on the DLBCL-1 dataset.} Shots refer to the number of annotations per class. Results are reported as mean $\pm$ standard deviation across all images. Bold values indicate the best performance across all baselines.}
\label{tab:few_shots}
\begin{tabular}{crrrr}
\toprule
\textbf{Shots} & \textbf{DINO-v2} & \textbf{UNI} & \textbf{CA-MAE} & \textbf{KRONOS} \\
\midrule
1  & 0.1336 $\pm$ 0.0566 & 0.1365 $\pm$ 0.0630 & 0.1363 $\pm$ 0.0564 & \textbf{0.3977 $\pm$ 0.1328} \\
5  & 0.1412 $\pm$ 0.0321 & 0.1463 $\pm$ 0.0351 & 0.1401 $\pm$ 0.0333 & \textbf{0.6523 $\pm$ 0.0845} \\
10 & 0.1443 $\pm$ 0.0268 & 0.1521 $\pm$ 0.0276 & 0.1408 $\pm$ 0.0261 & \textbf{0.7164 $\pm$ 0.0675} \\
20 & 0.1485 $\pm$ 0.0226 & 0.1628 $\pm$ 0.0244 & 0.1418 $\pm$ 0.0208 & \textbf{0.7534 $\pm$ 0.0581} \\
\bottomrule
\end{tabular}
\end{table}

\begin{table}[ht]
\centering
\caption{\textbf{Zero-shot cell phenotyping performance of KRONOS, DINO-v2, UNI, and CA-MAE in cHL, DLBCL-1 and DLBCL-2.} Performance is measured by adjusted rand index (ARI). Results are reported as mean $\pm$ standard deviation. Bold values indicate the best performance at each shot.}
\label{tab:zero_shot}
\begin{tabular}{lrrrr}
\toprule
Dataset & DINO-v2         & UNI             & CA-MAE          & KRONOS                   \\
\midrule
cHL     & 0.1723 $\pm$ 0.0147 & 0.1408 $\pm$ 0.0151 & 0.1323 $\pm$ 0.0210 & \textbf{0.3186 $\pm$ 0.0244} \\
DLBCL-1 & 0.0086 $\pm$ 0.0045 & 0.0165 $\pm$ 0.0093 & 0.0146 $\pm$ 0.0087 & \textbf{0.2589 $\pm$ 0.0344} \\
DLBCL-2 & 0.0057 $\pm$ 0.0010 & 0.0111 $\pm$ 0.0036 & 0.0090 $\pm$ 0.0030 & \textbf{0.2876 $\pm$ 0.0686} \\
\bottomrule
\end{tabular}
\end{table}

\begin{table}[ht]
\centering
\caption{\textbf{Comparison of KRONOS patch phenotyping performance against marker expression features in CLBCL-1.} Performance is stratified by ground-truth pixel ratio (i.e., proportion of the patch belonging to the majority phenotype). Results are reported as mean $\pm$ standard deviation.}
\label{tab:patch_phenotyping}
\begin{tabular}{clrr}
\toprule
\textbf{GT Pixels (\%)} & \textbf{Embedding Type} & \textbf{F1-Score} & \textbf{Balanced Accuracy} \\
\midrule
\multirow{2}{*}{20}  
  & Marker Expression & 0.4793 $\pm$ 0.0716 & 0.4704 $\pm$ 0.0469 \\
  & KRONOS Embedding  & \textbf{0.5569 $\pm$ 0.0698} & \textbf{0.5510 $\pm$ 0.0538} \\
\multirow{2}{*}{40}  
  & Marker Expression & 0.5014 $\pm$ 0.0768 & 0.4966 $\pm$ 0.0441 \\
  & KRONOS Embedding  & \textbf{0.5808 $\pm$ 0.0763} & \textbf{0.5800 $\pm$ 0.0539} \\
\multirow{2}{*}{60}  
  & Marker Expression & 0.5673 $\pm$ 0.0972 & 0.5841 $\pm$ 0.0344 \\
  & KRONOS Embedding  & \textbf{0.6406 $\pm$ 0.0948} & \textbf{0.6689 $\pm$ 0.0438} \\
\multirow{2}{*}{80}  
  & Marker Expression & 0.5992 $\pm$ 0.1115 & 0.6561 $\pm$ 0.0469 \\
  & KRONOS Embedding  & \textbf{0.6565 $\pm$ 0.1152} & \textbf{0.7143 $\pm$ 0.0459} \\
\multirow{2}{*}{100} 
  & Marker Expression & 0.5083 $\pm$ 0.1485 & 0.6885 $\pm$ 0.0814 \\
  & KRONOS Embedding  & \textbf{0.5671 $\pm$ 0.1413} & \textbf{0.7183 $\pm$ 0.0759} \\
\bottomrule
\end{tabular}
\end{table}

\begin{table}[ht]
\centering
\caption{\textbf{Comparison of KRONOS region classification against marker expression embeddings in prostate cancer subtyping.} Results are reported as mean $\pm$ standard deviation. Bold values indicate best performance per metric.}
\label{tab:region_classification}
\begin{tabular}{lrrrr}
\toprule
\textbf{Embedding Type} & \textbf{F1-Score} & \textbf{Balanced Accuracy} & \textbf{Average Precision} & \textbf{AUROC} \\
\midrule
Marker Expression & 0.5341 $\pm$ 0.1292 & 0.6007 $\pm$ 0.0731 & 0.7568 $\pm$ 0.0544 & 0.6664 $\pm$ 0.0668 \\
KRONOS Features   & \textbf{0.8326 $\pm$ 0.0527} & \textbf{0.8301 $\pm$ 0.0578} & \textbf{0.9413 $\pm$ 0.0091} & \textbf{0.9082 $\pm$ 0.0299} \\
\bottomrule
\end{tabular}
\end{table}

\begin{table}[ht]
\centering
\caption{\textbf{Comparison of KRONOS artifact detection against marker expression embeddings in DLBCL-1.} Results are reported as mean $\pm$ standard deviation. Bold values indicate best performance per metric.}
\label{tab:artifact_classification}
\begin{tabular}{lrrrr}
\toprule
\textbf{Embedding Type} & \textbf{F1-Score} & \textbf{Balanced Accuracy} & \textbf{Average Precision} & \textbf{AUROC} \\
\midrule
Marker Expression & 0.6536 $\pm$ 0.0207 & 0.6776 $\pm$ 0.0096 & 0.7416 $\pm$ 0.0210 & 0.8649 $\pm$ 0.0086 \\
KRONOS Features   & \textbf{0.9794 $\pm$ 0.0132} & \textbf{0.9797 $\pm$ 0.0147} & \textbf{0.9832 $\pm$ 0.0254} & \textbf{0.9911 $\pm$ 0.0137} \\
\bottomrule
\end{tabular}
\end{table}

\begin{table}[ht]
\centering
\caption{\textbf{Performance comparison between KRONOS, DINO-v2, UNI, and CA-MAE for K cluster distribution-based case classification in the CRC dataset.} Results reported as balanced accuracy with mean $\pm$ 95\% confidence interval across 100 runs. Bold values indicate best performance per cluster value.}
\label{tab:clustering_classification}
\begin{tabular}{crrrr}
\toprule
Cluster(K) & DINO-v2         & UNI             & CA-MEA          & KRONOS                   \\
\midrule
4          & 0.6532 $\pm$ 0.0164 & 0.6640 $\pm$ 0.0197 & 0.7103 $\pm$ 0.0151 & \textbf{0.7389 $\pm$ 0.0159} \\
8          & 0.6564 $\pm$ 0.0159 & 0.6665 $\pm$ 0.0191 & 0.6952 $\pm$ 0.0157 & \textbf{0.7584 $\pm$ 0.0162} \\
16         & 0.6816 $\pm$ 0.0169 & 0.6374 $\pm$ 0.0173 & 0.7059 $\pm$ 0.0153 & \textbf{0.7188 $\pm$ 0.0157} \\
\bottomrule
\end{tabular}
\end{table}

\begin{table}[ht]
\centering
\caption{\textbf{Performance comparison between KRONOS, DINO-v2, and UNI for patient stratification into responder and non-responder using two different embedding types.} Results are reported as mean $\pm$ 95\% confidence interval across 100 runs. Bold values indicate best performance per row.}
\label{tab:case_stratification}

\begin{tabular}{llrrr}
\toprule
Dataset & Feature type & DINO-v2         & UNI             & KRONOS                   \\
\midrule
\multirow{2}{*}{RCC}
& PCA-64       & 0.6655 $\pm$ 0.0149 & 0.6571 $\pm$ 0.0157 & \textbf{0.7672 $\pm$ 0.0132} \\
& PCA-256      & 0.6890 $\pm$ 0.0157 & 0.6723 $\pm$ 0.0158 & \textbf{0.7895 $\pm$ 0.0123} \\
\multirow{2}{*}{CTCL-B}
& PCA-64       & 0.5487 $\pm$ 0.0602 & 0.5750 $\pm$ 0.0576 & \textbf{0.6500 $\pm$ 0.0525} \\
& PCA-256      & 0.4750 $\pm$ 0.0576 & 0.4437 $\pm$ 0.0624 & \textbf{0.5863 $\pm$ 0.0593} \\
\multirow{2}{*}{CTCL-P}
& PCA-64       & 0.4400 $\pm$ 0.0850 & 0.3550 $\pm$ 0.0804 & \textbf{0.5900 $\pm$ 0.0793} \\
& PCA-256      & 0.4150 $\pm$ 0.0825 & 0.3050 $\pm$ 0.0749 & \textbf{0.5750 $\pm$ 0.0817} \\
\bottomrule
\end{tabular}
\end{table}

\begin{table}[ht]
\centering
\caption{\textbf{Performance comparison for cell retrieval between KRONOS, DINO-v2, UNI, and CA-MAE.} Results reported as top-k accuracy with mean $\pm$ standard deviation. Bold values indicate best performance per Top-k.}
\label{tab:cell_retrieval}
\begin{tabular}{lcrrrr}
\toprule
Datasets & Top-k & DINO-v2         & UNI             & CA-MAE          & KRONOS                   \\
\midrule
\multirow{3}{*}{cHL}
& 1     & 0.3217 $\pm$ 0.0165 & 0.2556 $\pm$ 0.0099 & 0.2530 $\pm$ 0.0131 & \textbf{0.5065 $\pm$ 0.0086} \\
& 3     & 0.5706 $\pm$ 0.0187 & 0.4899 $\pm$ 0.0103 & 0.4826 $\pm$ 0.0135 & \textbf{0.7214 $\pm$ 0.0140} \\
& 5     & 0.6898 $\pm$ 0.0144 & 0.6150 $\pm$ 0.0097 & 0.6070 $\pm$ 0.0129 & \textbf{0.8004 $\pm$ 0.0168} \\
\multirow{3}{*}{DLBCL-1}
& 1     & 0.1170 $\pm$ 0.0029 & 0.1218 $\pm$ 0.0031 & 0.1142 $\pm$ 0.0031 & \textbf{0.6502 $\pm$ 0.0340} \\
& 3     & 0.3100 $\pm$ 0.0059 & 0.3179 $\pm$ 0.0074 & 0.3038 $\pm$ 0.0051 & \textbf{0.8151 $\pm$ 0.0135} \\
& 5     & 0.4587 $\pm$ 0.0085 & 0.4667 $\pm$ 0.0096 & 0.4511 $\pm$ 0.0078 & \textbf{0.8697 $\pm$ 0.0078} \\
\multirow{3}{*}{DLBCL-2}
& 1     & 0.1206 $\pm$ 0.0014 & 0.1243 $\pm$ 0.0014 & 0.1150 $\pm$ 0.0010 & \textbf{0.6780 $\pm$ 0.0637} \\
& 3     & 0.3168 $\pm$ 0.0022 & 0.3240 $\pm$ 0.0026 & 0.3068 $\pm$ 0.0024 & \textbf{0.8293 $\pm$ 0.0469} \\
& 5     & 0.4668 $\pm$ 0.0025 & 0.4750 $\pm$ 0.0041 & 0.4558 $\pm$ 0.0038 & \textbf{0.8757 $\pm$ 0.0393} \\
\bottomrule
\end{tabular}
\end{table}

\begin{table}[ht]
\centering
\caption{\textbf{Performance comparison for patch retrieval between KRONOS, DINO-v2, UNI, and CA-MAE using root mean squared error metric.} Results are reported as mean $\pm$ standard deviation. Bold values indicate best performance per Top-k.}
\label{tab:patch_retrieval}
\begin{tabular}{lcrrrr}
\toprule
Datasets & Top-k & DINO-v2          & UNI              & CA-MAE           & KRONOS           \\
\midrule
\multirow{3}{*}{DLBCL-1}  
& 1     & 15.3894 $\pm$ 2.9032 & 13.3446 $\pm$ 1.1263 & 16.5907 $\pm$ 2.4181 & \textbf{10.7976 $\pm$ 1.8082} \\
& 3     & 11.6337 $\pm$ 2.2880 & 10.3069 $\pm$ 0.8545 & 12.7466 $\pm$ 1.9419 & \textbf{9.1376 $\pm$ 1.9213}  \\
& 5     & 10.2334 $\pm$ 1.8976 & 9.2038 $\pm$ 0.8679  & 11.4334 $\pm$ 1.7588 & \textbf{8.4215 $\pm$ 1.8221}  \\
\multirow{3}{*}{DLBCL-2} 
& 1     & 15.6984 $\pm$ 1.3068 & 14.4233 $\pm$ 1.5871 & 18.4322 $\pm$ 0.9948 & \textbf{10.2071 $\pm$ 0.9040} \\
& 3     & 12.2468 $\pm$ 1.3281 & 11.3512 $\pm$ 1.6390 & 14.2778 $\pm$ 0.6687 & \textbf{8.5496 $\pm$ 0.8622}  \\
& 5     & 11.0626 $\pm$ 1.3344 & 10.2383 $\pm$ 1.6699 & 12.7449 $\pm$ 0.5763 & \textbf{7.8742 $\pm$ 0.8755} \\
\bottomrule
\end{tabular}
\end{table}

\begin{table}[ht]
\centering
\caption{\textbf{Performance comparison for case retrieval between KRONOS, DINO-v2, UNI, and CA-MAE.} Results are reported as mean $\pm$ 95\% confidence interval. Bold values indicate best performance per metric.}
\label{tab:case_retrieval}
\begin{tabular}{lrrrr}
\toprule
Datasets          & DINO-v2         & UNI             & CA-MEA          & KRONOS                   \\
\midrule
Accuracy          & 0.7529 $\pm$ 0.0160 & 0.7436 $\pm$ 0.0189 & 0.7262 $\pm$ 0.0195 & \textbf{0.8233 $\pm$ 0.0097} \\
F1-Score          & 0.7402 $\pm$ 0.0174 & 0.7385 $\pm$ 0.0194 & 0.7115 $\pm$ 0.0211 & \textbf{0.8190 $\pm$ 0.0103} \\
Balanced Accuracy & 0.7349 $\pm$ 0.0163 & 0.7356 $\pm$ 0.0177 & 0.7032 $\pm$ 0.0194 & \textbf{0.8076 $\pm$ 0.0103} \\
\bottomrule
\end{tabular}
\end{table}

\begin{table}[]
\centering
\caption{\textbf{Hyperparameters used for KRONOS pretraining with DINO-v2.}}
\label{tab:kronos_config}
\begin{tabular}{ll}
\toprule
\textbf{Hyperparameter}                    & \textbf{Value}            \\
\midrule
Layers                            & 12               \\
Heads                             & 6                \\
Registers                         & 16               \\
Patch size                        & 224x224          \\
Head activation                   &                  \\
Embedding dimension               & 384              \\
Drop path rate                    & 0.3              \\
Global crop scale                 & {[}0.48, 1{]}    \\
Global crop number                & 2                \\
Local crop scale                  & {[}0.16, 0.48{]} \\
Local crop number                 & 8                \\
Gradient clipping max norm        & 3                \\
Shared head                       & Yes              \\
AdamW Beta                        & {[}0.9, 0.999{]} \\
Batch size                        & 1024             \\
Freeze last layer epochs          & 1                \\
Warmup epochs                     & 10               \\
Warmup teacher temperature epochs & 30               \\
Max epochs                        & 100              \\
Learning rate schedule            & Cosine           \\
Learning rate (start)             & 0.004            \\
Learning rate (post warmup)       & 0.004            \\
Learning rate (final)             & 1.00E-06         \\
Teacher temperature (start)       & 0.04             \\
Teacher temperature (final)       & 0.07             \\
Teacher momentum (start)          & 0.992            \\
Teacher momentum (final)          & 1                \\
Weight decay (start)              & 0.04             \\
Weight decay (end)                & 0.4              \\
Automatic mixed precision         & fp16            \\
\bottomrule
\end{tabular}
\end{table}

\end{document}